\documentclass[conference]{IEEEtran}
\IEEEoverridecommandlockouts
\usepackage{cite}
\usepackage{amsmath,amssymb,amsfonts}
\usepackage{graphicx}
\usepackage{textcomp}
\usepackage{xcolor}
\usepackage{hyperref}
\usepackage{multirow}
\usepackage{multicol}
\usepackage{threeparttable}
\usepackage{tabularx,booktabs}  
\usepackage{boldline}
\usepackage{subfig}
\usepackage{wrapfig}
\usepackage{amsmath}
\usepackage{algorithm}
\usepackage{algpseudocode}
\usepackage{flushend}
\usepackage{tikz}
\newcommand\copyrighttext{%
  \footnotesize $\copyright$ 2022 IEEE. Personal use of this material is permitted. Permission from IEEE must be obtained for all other uses, in any current or future media, including reprinting/republishing this material for advertising or promotional purposes, creating new collective works, for resale or redistribution to servers or lists, or reuse of any copyrighted component of this work in other works.}
\newcommand\copyrightnotice{%
\begin{tikzpicture}[remember picture,overlay]
\node[anchor=south,yshift=10pt] at (current page.south) {\fbox{\parbox{\dimexpr\textwidth-\fboxsep-\fboxrule\relax}{\copyrighttext}}};
\end{tikzpicture}%
}
\usepackage{xcolor}
\newcommand*\circled[1]{\tikz[baseline=(char.base)]{
            \node[shape=circle,fill,inner sep=0.5pt] (char) {\textcolor{white}{#1}};}}

\def\BibTeX{{\rm B\kern-.05em{\sc i\kern-.025em b}\kern-.08em
    T\kern-.1667em\lower.7ex\hbox{E}\kern-.125emX}}
\begin{document}

\title{Demand Layering for Real-Time DNN Inference with Minimized Memory Usage}

\author{\IEEEauthorblockN{Mingoo Ji${}^{1, 2}$, Saehanseul Yi${}^{3}$, Changjin Koo${}^{1}$, Sol Ahn${}^{1}$, Dongjoo Seo${}^{3}$, Nikil Dutt${}^{3}$, and Jong-Chan Kim${}^{1,4}$}
\IEEEauthorblockA{
{${}^{1}$Graduate School of Automotive Engineering, Kookmin University, Korea}\\ 
{${}^{2}$Automotive Electronics Advanced Development TFT, Hyundai Mobis, Korea}\\ 
{${}^{3}$Department of Computer Science, University of California, Irvine, USA}\\
{${}^{4}$Department of Automobile and IT Convergence, Kookmin University, Korea}\\
{\small Correspondence: jongchank@kookmin.ac.kr}
}}

\maketitle

\begin{abstract}
When executing a deep neural network (DNN), its model parameters are loaded into GPU memory before execution, incurring a significant GPU memory burden. There are studies that reduce GPU memory usage by exploiting CPU memory as a swap device. However, this approach is not applicable in most embedded systems with integrated GPUs where CPU and GPU share a common memory. In this regard, we present Demand Layering, which employs a fast solid-state drive (SSD) as a co-running partner of a GPU and exploits the layer-by-layer execution of DNNs. In our approach, a DNN is loaded and executed in a layer-by-layer manner, minimizing the memory usage to the order of a single layer. Also, we developed a pipeline architecture that hides most additional delays caused by the interleaved parameter loadings alongside layer executions. Our implementation shows a 96.5\% memory reduction with just 14.8\% delay overhead on average for representative DNNs. Furthermore, by exploiting the memory-delay tradeoff, near-zero delay overhead (under 1~ms) can be achieved with a slightly increased memory usage (still an 88.4\% reduction), showing the great potential of Demand Layering.
\end{abstract}

\copyrightnotice
\section{Introduction} \label{sec:intro}

To enable efficient deep neural network (DNN) inference with low-cost embedded hardware, its memory requirement should be minimized. For that, a typical approach is to apply pruning and quantization~\cite{han2015deep, he2018amc, liu2017learning} that reduce the number of model parameters, however, at the cost of unavoidable accuracy loss. Once a model is fixed, all the parameters are loaded into system memory before execution. To the best of our knowledge, most state-of-the-art DNN frameworks employ this method despite its excessive memory usage. However, in the era of large-scale models~\cite{goodfellow2016deep, huang2019gpipe, wang2019supporting, park2020hetpipe, bernstein2021freely} and concurrent DNNs~\cite{bai2020pipeswitch, kang2021lalarand, xiang2019pipelined}, we argue that this na\"ive approach is no longer viable, and thus a new system approach is needed that can alleviate this excessive memory requirement.

Recent studies try to reduce the memory usage of DNN inference by efficiently managing activation buffers between DNN layers~\cite{pisarchyk2020efficient, lin2020mcunet, lin2021mcunetv2}. However, they are not applicable for storing model parameters. Besides, SwapAdvisor~\cite{huang2020swapadvisor} provides a general method by utilizing inexpensive CPU memory as a swap device of scarce GPU memory. This method is promising in discrete GPU (dGPU) systems with separate CPU and GPU memory. However, most embedded systems use integrated GPUs (iGPUs), where CPU and GPU share a common memory system~\cite{bateni2020co}. In such systems, reducing GPU memory at the cost of increased CPU memory does not provide any benefit.

\begin{figure}
    \centering
    \includegraphics[width=0.35\textwidth]{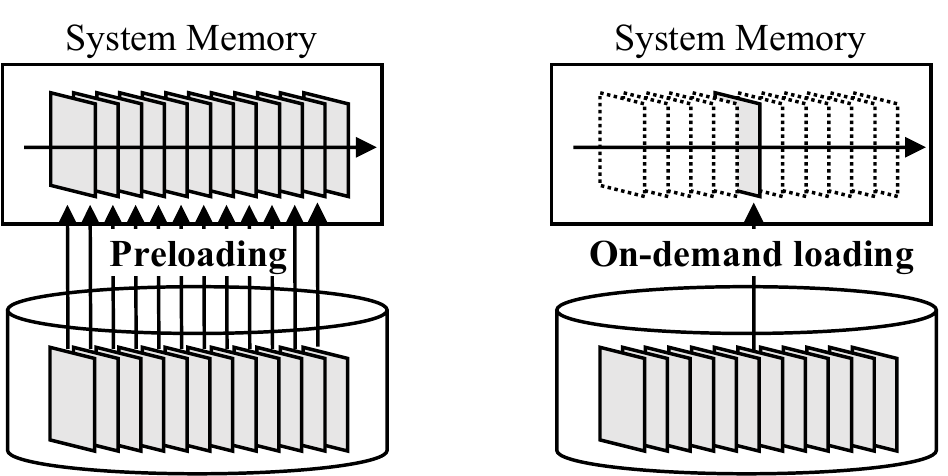}
    \caption{Preloading vs. Demand Layering.}
    \label{fig:concept}
    \vspace{-0.3cm}
\end{figure}

With this motivation, this study aims to reduce the memory usage of iGPU-based DNN inference systems, explicitly targeting the memory for model parameters. Our idea is to borrow the concept of {\em demand paging} in conventional operating systems, where program instructions are loaded to CPU memory on demand in the granularity of {\em pages} (typically sized 4 KB - 16 KB). Similarly, exploiting the layer-by-layer execution of DNNs, we propose {\em Demand Layering} that loads model parameters on demand in the granularity of {\em layers} while dropping previous layers of no use. In this manner, the memory requirement is significantly reduced to the order of a single layer from the order of the entire model. Fig.~\ref{fig:concept} highlights the difference between the preloading architecture and our Demand Layering.

However, the memory reduction is not free. It comes at the cost of increased delays. Thus, we conducted a thorough delay analysis, which found out that the inference delay can be analyzed in terms of the following three operations in it:
\begin{itemize}
    \item {\bf Read:} Model parameters are read into CPU memory.
    \item {\bf Copy:} Model parameters are copied to GPU memory.
    \item {\bf Kernel:} DNN layers are executed by GPU kernels.
\end{itemize}

In the preloading architecture, all the read and copy operations are only in the {\em initialization phase}; thus, its inference delay is just the sum of GPU kernel executions. In contrast, Demand Layering repeatedly conducts read and copy operations during the {\em inference phase}, which potentially causes extra delays. For that, our baseline approach is to employ a high-performance solid-state drive (SSD). Compared with eMMC storages typically with 300~MB/s sequential read performance, M.2 NVMe SSDs provide up to 7000~MB/s of sequential read performance~\cite{huang2022ssds}. Although random reads are somewhat slower, most DNN model files exhibit sequential access patterns due to the inherent sequential nature of DNN executions~\cite{kang2021lalarand, xiang2019pipelined}.

Even with the fastest SSD, extra delays are still significant. Thus, our next approach is to hide away the delays as much as possible by pipelined execution of read, copy, and kernel operations.
Fortunately, even in iGPU systems, these three operations can run in parallel, because read operations can be carried out by CPU while copy and kernel operations are being processed by GPU. Even better, Nvidia GPUs have two separate processing units: a copy engine (CE) and an execution engine (EE). The CE can process copy operations while the EE is executing GPU kernels~\cite{amert2017gpu, olmedo2020dissecting}. As a result, read, copy, and kernel operations can run fully in parallel. Based on this parallel hardware architecture, we developed and evaluated a number of software pipeline architectures on an Nvidia Jetson AGX Xavier platform with various DNNs. The remainder of this section introduces the case with YOLOv4~\cite{bochkovskiy2020yolov4} in particular, whose model size is 245.8~MB and its average inference delay in the preloading architecture is 160.8~ms. Besides, the largest layer size is 18.0~MB.

{\bf Synchronous pipeline.} In the 3-stage synchronous pipeline architecture, its read, copy, and kernel stages advance while synchronized with a common pipeline cycle. Since kernel operations are usually the longest among the three stages, most read and copy operations are hidden behind kernel operations. This pipeline architecture needs two inter-stage buffers: (i) a CPU memory buffer between read and copy stages and (ii) a GPU memory buffer between copy and kernel stages. Since each buffer needs to hold just the layer being processed, the required buffer size is the size of the largest layer.
In addition, the buffers should be {\em double-buffered} because, for example, a read to the CPU buffer can happen simultaneously with a copy from the same buffer. The same applies to the GPU buffer. Our implementation provides an 85.4\% memory reduction (to 72.0~MB) with 23.7\% delay overhead (to 198.9~ms).

{\bf Asynchronous pipeline.} If a read operation happens to be the longest in a synchronous pipeline cycle, it causes a GPU idling interval, negatively impacting the delay.
To minimize such unwanted delays, our architecture is modified to an asynchronous pipeline, where pipeline stages advance at their own paces~\cite{nowick2011high}. Between the pipeline stages, we introduce two {\em circular buffers} that can barely hold the largest layer each, instead of the two pairs of double buffers used in the synchronous architecture, cutting the memory requirement in half. Our implementation provides a 92.7\% memory reduction (to 36.0~MB) with 12.7\% delay overhead (to 181.2~ms).

{\bf Two-stage pipeline.} Recent iGPU-based system on chips (SoCs) (e.g., Nvidia Xavier) provide a special memory management scheme so that a memory buffer can be accessed both from CPU and GPU~\cite{bateni2020co}. This zero-copy memory eliminates the need of copy operations, enabling a 2-stage pipeline. By this architecture, the memory requirement is further reduced to just the order of a single layer. Our implementation provides a 96.3\% memory reduction (to 18.0~MB) with 21.5\% delay overhead (to 195.3~ms).

{\bf Memory-delay tradeoff.} In the asynchronous pipeline architectures, we can intentionally increase the circular buffer size to exploit the tradeoff relation between memory and delay. Thus, we can devise an iterative optimization process by gradually increasing the buffer size until there is no further delay reduction. By this optimization method, we can find the {\em minimal delay} configuration. As a result, near-zero ($<$ 1.0~ms) delay overhead is achieved by a slight increase in memory usage (from 18.0~MB to 52.8~MB).

The contributions of this study can be summarized as:
\begin{itemize}
    \item We propose Demand Layering for minimized memory usage in DNN inference systems by loading and executing layers in a layer-by-layer manner.
    \item Three pipeline architectures are presented that minimize the extra delay overhead of Demand Layering.
    \item The pipeline architectures are implemented and evaluated on Nvidia Jetson AGX Xavier, showing significant memory reductions with near-zero delay overhead.
\end{itemize}

\section{Preliminaries} \label{sec:prelim}

\subsection{Deep Neural Networks (DNNs)} \label{sec:dnn}

In contrast to conventional programs, which are sequences of {\em instructions}, DNNs are sequences of {\em parameters}, organized by layers such as convolutional and fully connected layers. The parameters are produced in a training phase and stored in a DNN {\em model file}, whose file format depends on the DNN framework of your choice. For example, Darknet~\cite{darknet} uses .weights binary files. PyTorch~\cite{paszke2019pytorch} uses .pt or .pth files, which are serialized binary files by the Python pickle module. TensorFlow~\cite{abadi2016tensorflow} uses .pb files, which are binary files by the ProtoBuf format.

Regardless of the file format, the model files must be loaded to GPU memory in the initialization phase. Then, in the inference phase, the preloaded parameters are interpreted and executed by a DNN inference framework in a {\em layer-by-layer} manner~\cite{kang2021lalarand, bai2020pipeswitch}. This preloading architecture inherently imposes a significant GPU memory burden for storing the entire model parameters, especially serious in multi-DNN systems.

\subsection{Integrated CPU-GPU Systems} \label{sec:integrated}

\begin{figure}
    \centering
    \includegraphics[width=0.45\textwidth]{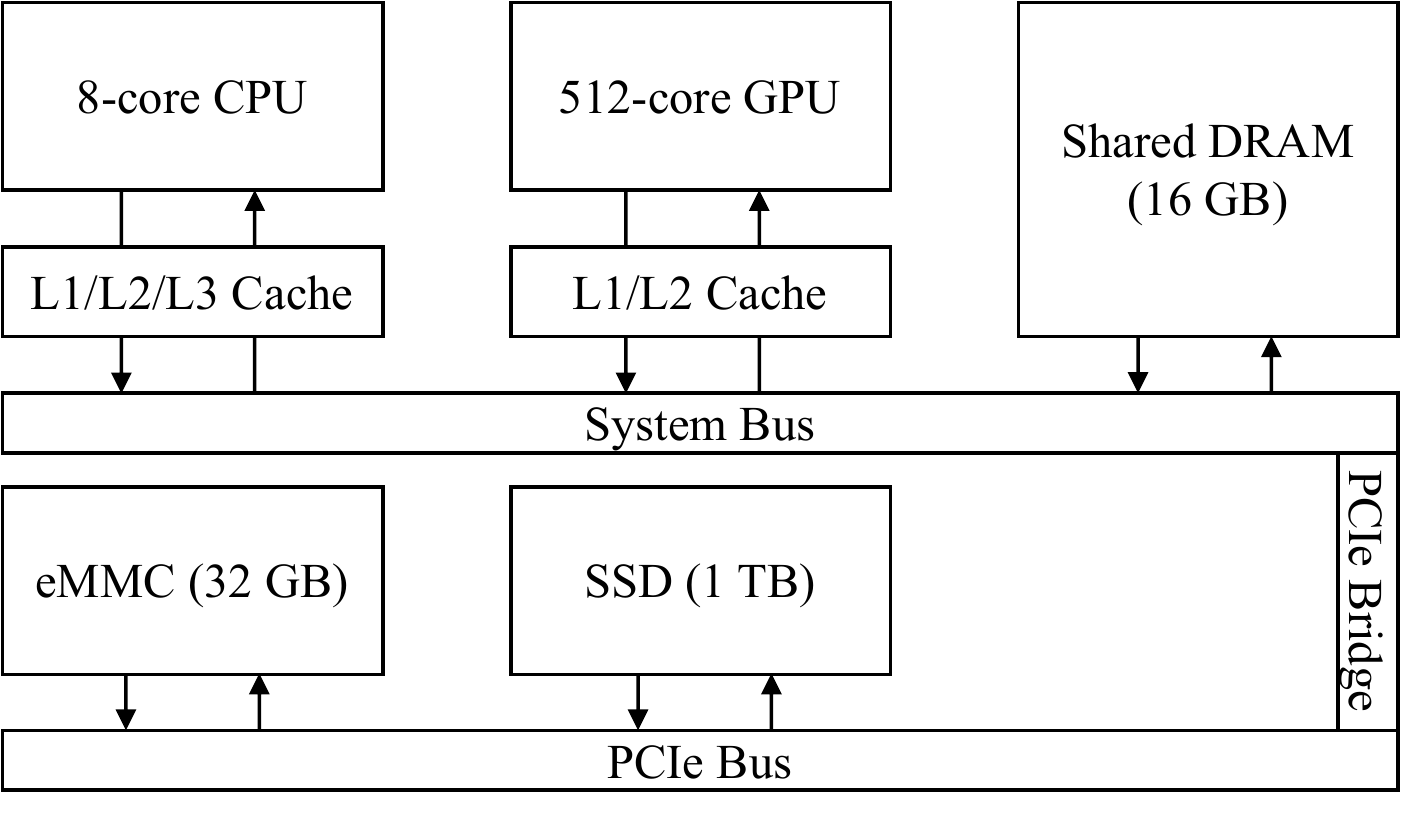}
    \caption{Nvidia Jetson AGX Xavier hardware architecture.}
    \label{fig:xavier}
    \vspace{0.11cm}
\end{figure}

When designing embedded systems for DNN applications, iGPUs are highly preferred to dGPUs due to the advantage in its size, weight, and power (SWaP) properties~\cite{bateni2020co}. In contrast to dGPUs, iGPUs share the same physical memory space with CPU. 
In such systems, GPU memory optimization at the expense of CPU memory cannot make a beneficial deal. Instead, a holistic CPU-GPU memory optimization method is required.

A typical example of integrated CPU-GPU systems is Nvidia Jetson AGX Xavier, which is our experimental platform. Fig.~\ref{fig:xavier} shows its internal architecture with 16~GB shared DRAM, an 8-core 64-bit ARM CPU, and a 512-core integrated Volta GPU connected through a system bus. Additionally, it is equipped with an M.2 NVMe interface through a PCI express (PCIe) bus that can host an optional SSD besides its built-in 32~GB eMMC storage.

\subsection{Solid-State Drives (SSDs)} \label{sec:ssd}

For many years, eMMC storages have dominated most embedded systems since conventional embedded applications did not require either TB-scale capacity or GB/s-level bandwidth. However, in recent data-intensive applications like autonomous driving, a vast amount of data should be stored and retrieved in real time, requiring huge storage capacities and high bandwidth. Since neither of them can be achieved by eMMC devices, a viable alternative is to employ SSDs in such data-centric embedded systems. Recent commercial off-the-shelf (COTS) SSDs can satisfy such excessive requirements with their ever-growing capacity and bandwidth.

Our experimental platform is also equipped with a Samsung 980 PRO NVMe M.2 SSD with 1~TB capacity and its officially announced 7000~MB/s sequential read performance. The SSD is connected to both CPU and GPU via a PCIe Gen4 interface. In our target application (i.e., DNN inference), the SSD is used to store DNN model files, which are usually above hundreds of megabytes. Furthermore, in multi-DNN systems, the storage requirement is far more significant, making SSDs an ideal choice for storing DNN model files.

\begin{figure}
    \centering
    \includegraphics[width=0.45\textwidth]{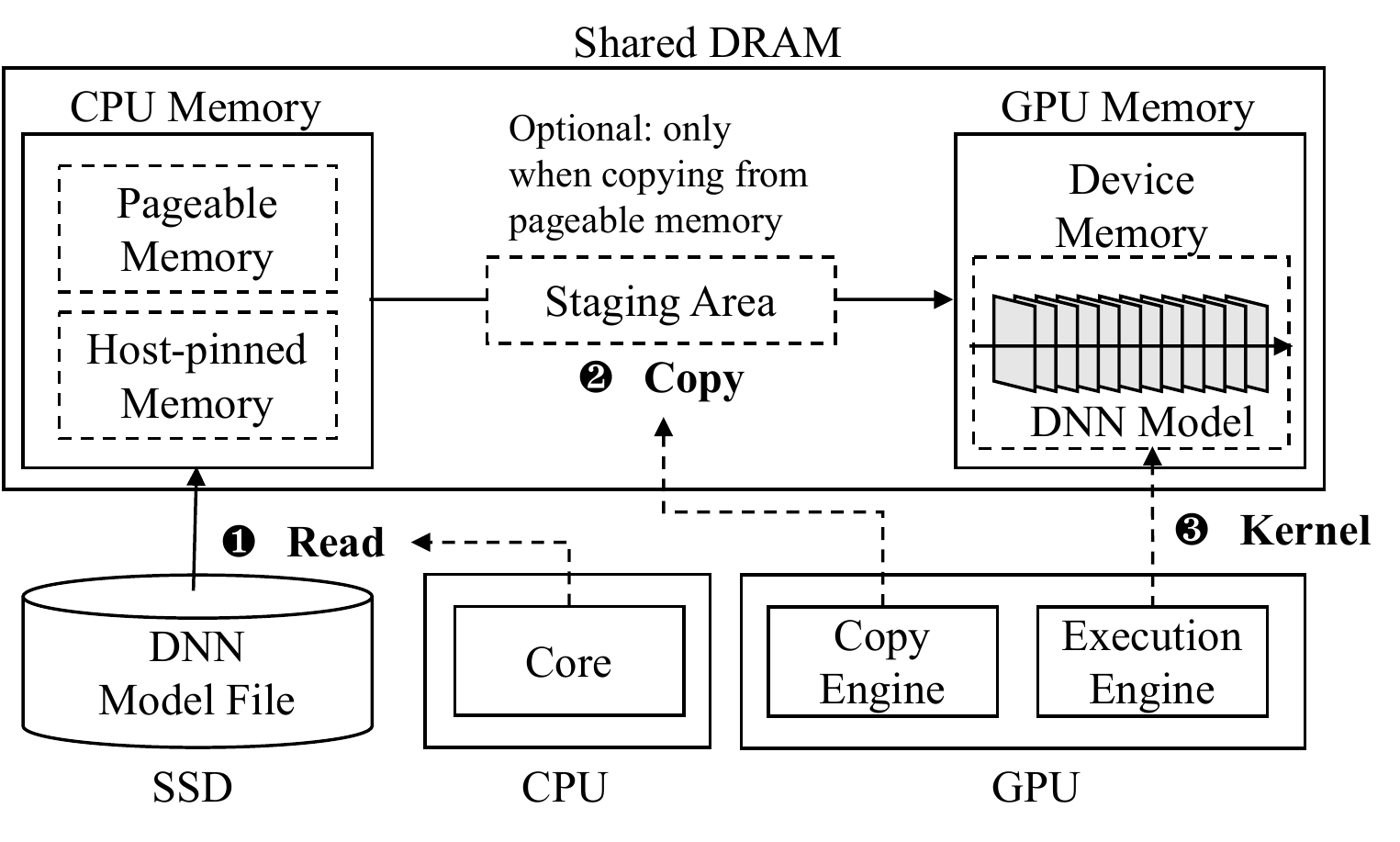}
    \caption{Data flow of model parameters.}
    \label{fig:flow}
\end{figure}

\begin{figure*}
    \centering 
    \subfloat{\label{fig:layer_time}\includegraphics[width=1\textwidth]{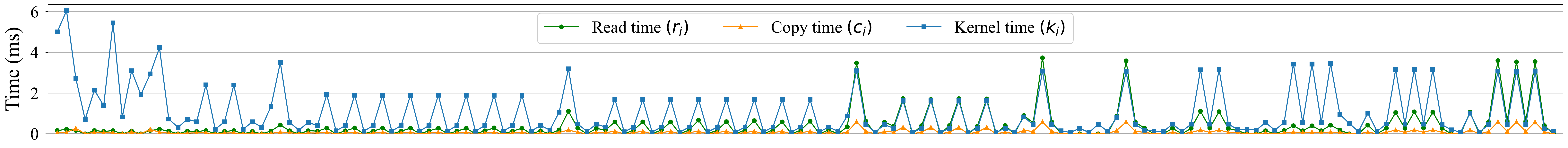}}\\
    \vspace{-0.3cm}
    \subfloat{\label{fig:layer_size}\includegraphics[width=1\textwidth]{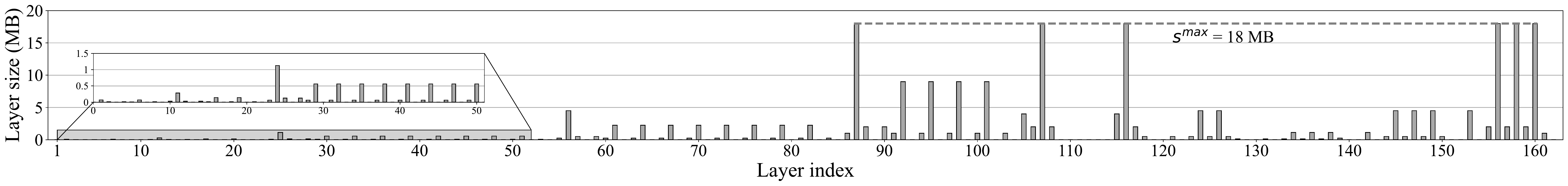}}\\
\caption{Layer-wise measurements of the YOLOv4 object detection network (input resolution = 608 $\times$ 608).}
\label{fig:layer}

\end{figure*}

\subsection{Data Flow of DNN Model Parameters} \label{sec:flow}

To begin an inference (i.e., forward propagation) on a DNN model, the entire parameters should be in GPU memory such that GPU kernels can directly access them. For that, a three-step approach is usually used, which is depicted in Fig.~\ref{fig:flow}. The model file is first read from disk to a CPU memory buffer (\circled{\raisebox{-.9pt} 1}). 
When allocating CPU buffers, there are several choices provided by the Nvidia CUDA runtime, which will be detailed in Section~\ref{sec:copy}.
Then the parameters are copied to a GPU memory buffer (\circled{\raisebox{-.9pt} 2}). 
When the source CPU buffer happens to be a pageable memory by the usual \texttt{malloc()} function that is not under the control of the CUDA runtime, the copy is done via a hidden staging area, incurring possible blockings and delays in case of a staging area shortage. After the copy operation, GPU kernels can access and execute the DNN layers in the GPU memory buffer (\circled{\raisebox{-.9pt} 3}). As explained in Section~\ref{sec:integrated}, CPU and GPU memory buffers are from the same shared DRAM space. Thus, both buffers should be accounted for when estimating the memory usage of a DNN inference system. The read operation is processed by CPU, while the copy and kernel operations are executed by GPU. Since GPUs have two separate processing units for them (i.e., {\em copy engine} and {\em execution engine}), read and copy operations can run simultaneously~\cite{yang2018avoiding}. As a result, read, copy, and kernel operations can run fully in parallel, providing a great chance for optimizing the DNN execution architecture. 

\subsection{Observations and Our Motivation} \label{sec: motivation}

Meanwhile, we have the following observations during the investigation on various DNN inference frameworks in GPU-based embedded systems:\\
{\bf (i) Memory burden in DNN inference.} To the best of our knowledge, most DNN inference frameworks preload the whole model parameters from disk to memory in the initialization phase to avoid disk operations in the inference phase. However, this {\em preloading} architecture permanently occupies a significant amount of system memory for storing model parameters, which is not acceptable in resource-constrained embedded systems.\\
{\bf (ii) Layer-by-layer DNN execution:} DNN models have a layered structure, where there are strict data dependencies between layers. Model files are also organized following the layered architecture. Most notably, when a certain layer is executing, only that layer's parameters are accessed, and the rest of the parameters are irrelevant to the current layer execution.\\
{\bf (iii) High-performance SSDs.} Most recent SSDs are fast enough, reaching the speed of 7000~MB/s for sequential reads. Certainly, the speed of random reads is far slower than sequential reads. Fortunately, however, what we need for the model file is only sequential reads that can best extract the peak performance of SSDs.

{\bf Motivation.} With the above observations, our intuition is that the memory usage can be drastically reduced by loading and unloading model parameters by a layer's granularity in the inference phase without preloading them in the initialization phase. By that, the maximum memory usage will be reduced from the order of the entire model to the order of a single layer. However, read and copy operations are additionally performed in the inference phase, potentially adding extra delays if they are not adequately overlapped with kernel executions using pipeline architectures.

{\bf Initial Profiling.} To pre-inspect the optimization opportunity, we measured the layer execution time of the popular YOLOv4 object detection model using the Darknet framework on an Nvidia Jetson AGX Xavier platform.
The upper graph in Fig.~\ref{fig:layer} shows each layer's average read, copy, and kernel execution times for 100 inference iterations, while the lower graph shows the size of each corresponding layer. Comparing the two graphs gives a hint of the strong correlation between timings and layer sizes.
Fig.~\ref{fig:hist} shows the measured distributions, indicating that the read and copy times are significantly shorter than the kernel times in most layers, showing the potential for overlapping read and copy operations behind kernel executions.

\section{System Model and Problem Description} \label{sec:model}

\subsection{System Model}

We assume an integrated CPU-GPU system equipped with an SSD, where a DNN inference engine runs a given DNN with $N$ layers, denoted by $\{l_1, l_2, \cdots, l_N\}$. The whole model parameters are stored in a model file in the SSD, which is sequentially organized by layers. Each $i$-th layer $l_i$'s size is denoted by $s_i$, while the largest layer size is denoted by $s^{max}$. In the timing perspective, each layer is characterized by a tuple $l_i$ = $(r_i, c_i, k_i)$, where $r_i$ is the time for reading, $c_i$ is the time for copying, and $k_i$ is the time for kernel execution. Note that $r_i$, $c_i$, $k_i$, and $s_i$ are random variables, not fixed values like worst-case or average execution times. For ease of presentation, we rather use {\em read time}, {\em copy time}, and {\em kernel time} when referring to the above layer-wise timing properties. Also, when specifically referring to the read, copy, and kernel operations themselves, we use capital letters ($R_i$, $C_i$, $K_i$) instead.

\begin{figure}
    \centering 
    \subfloat[Read time distribution.]
    {\label{fig:read}\includegraphics[width=0.5\linewidth]{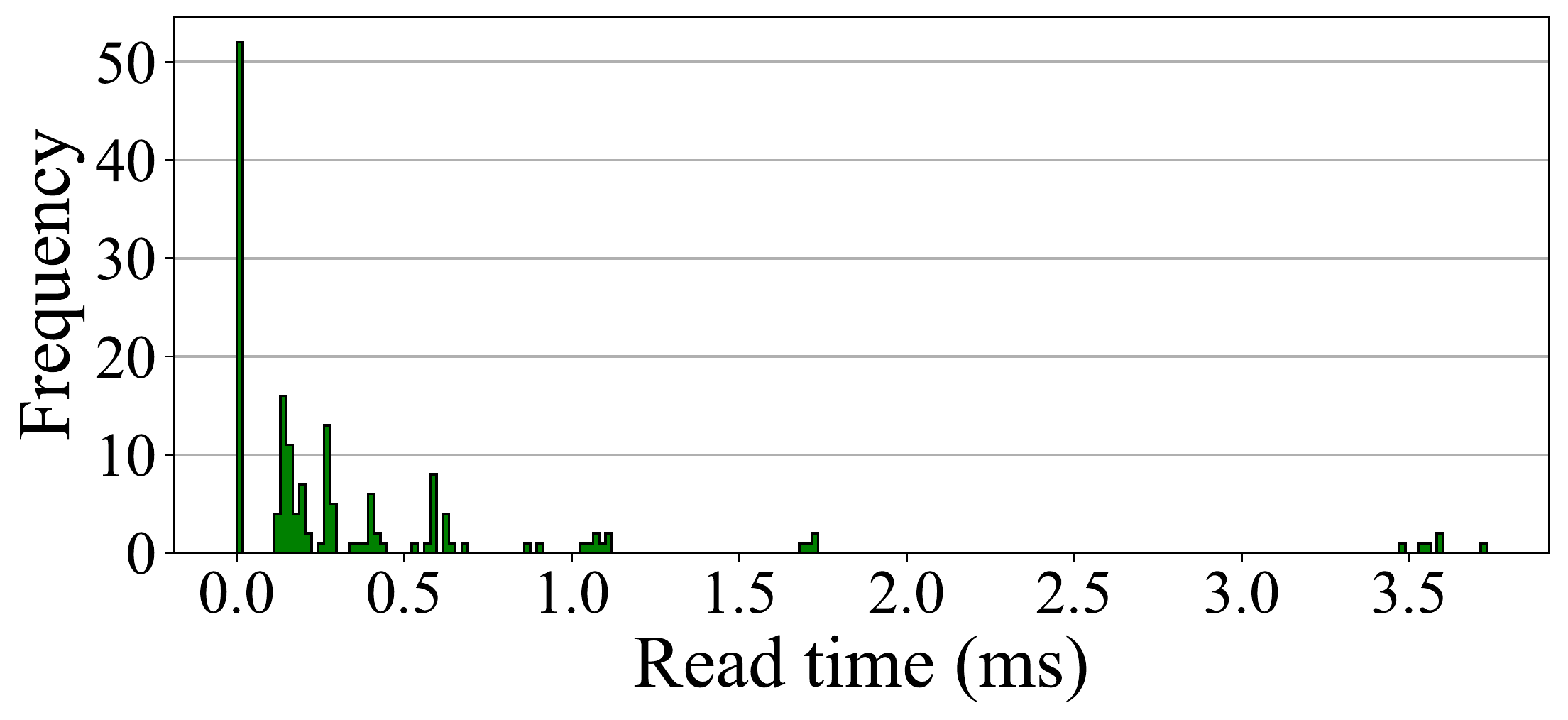}}
    \subfloat[Copy time distribution.]
    {\label{fig:copy}\includegraphics[width=0.5\linewidth]{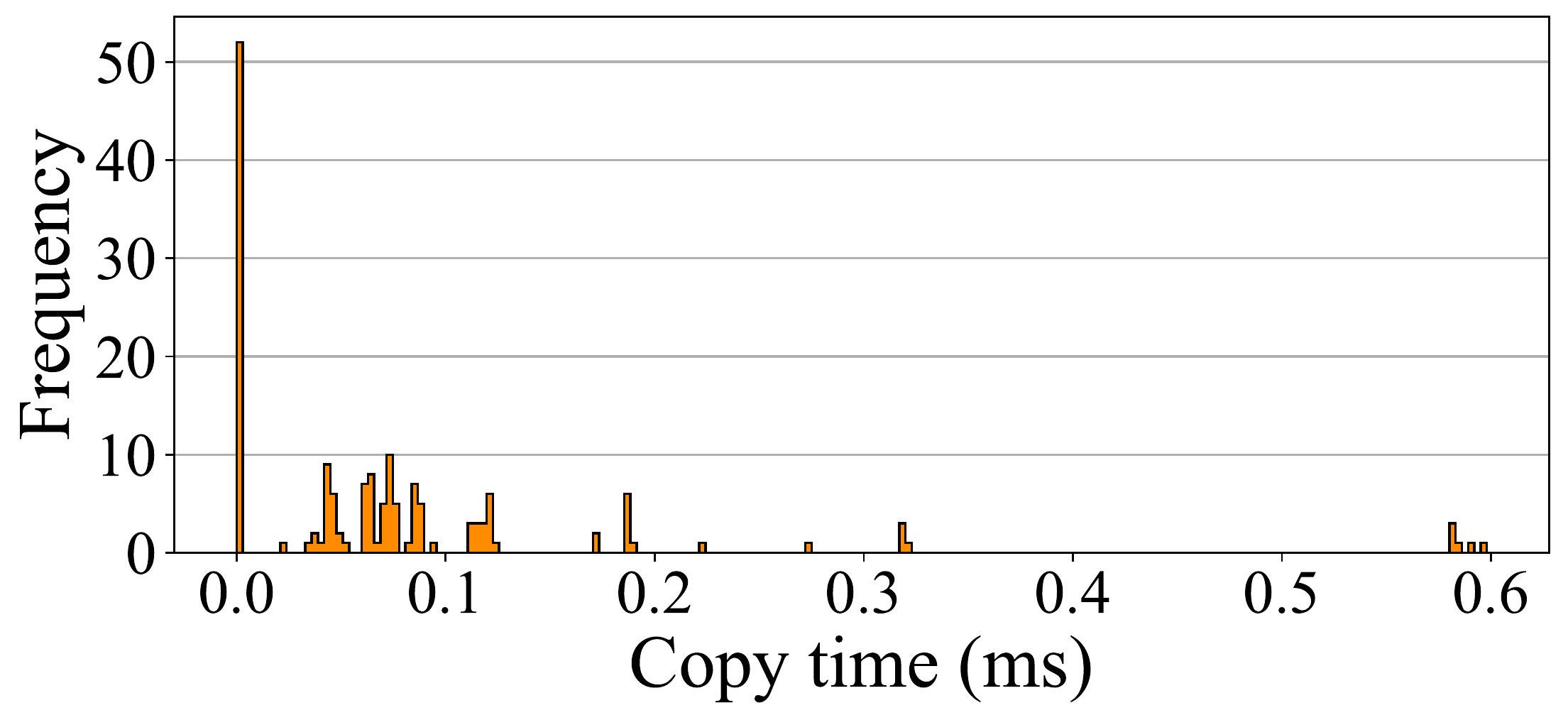}}\\
    \subfloat[Kernel time distribution.]
    {\label{fig:kernel}\includegraphics[width=0.5\linewidth]{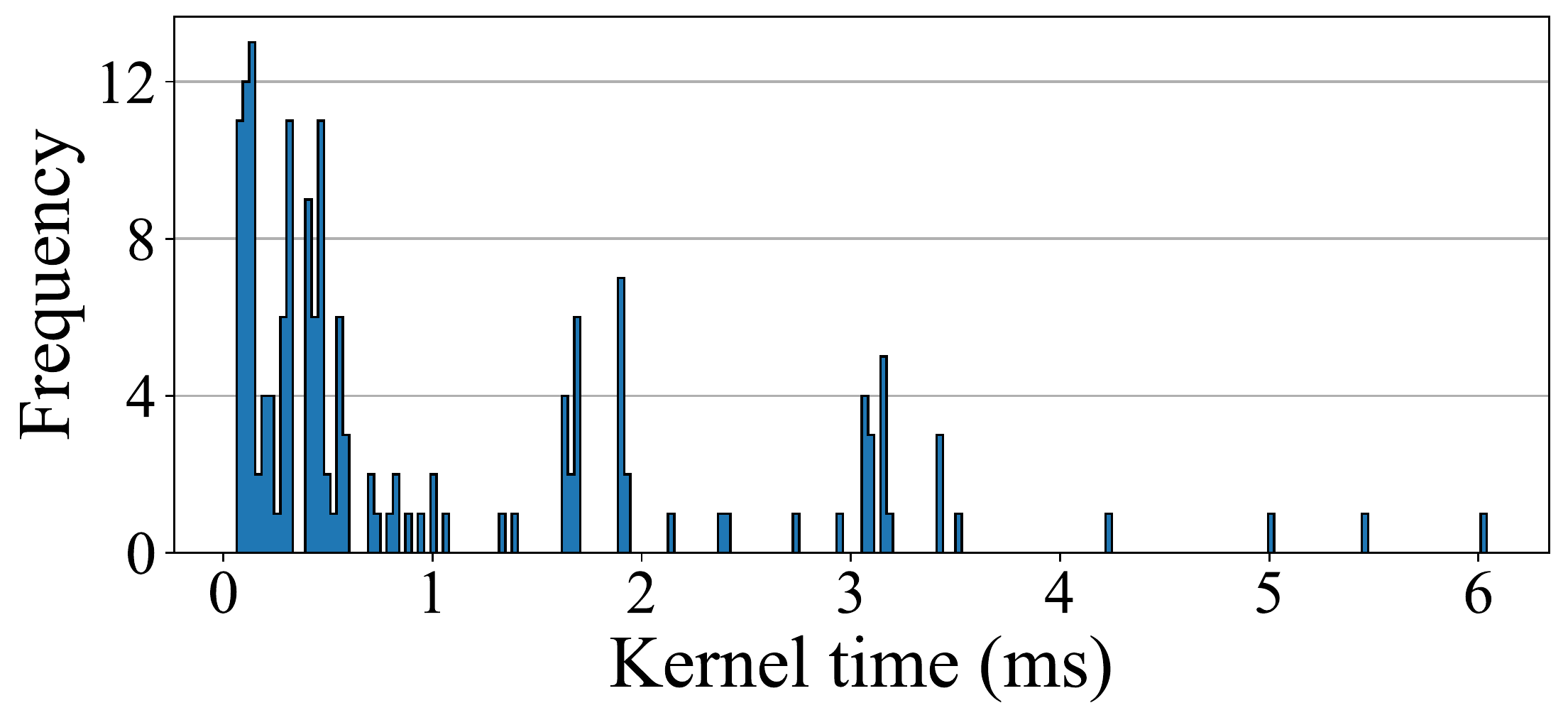}}
    \subfloat[Layer size distribution.]
    {\label{fig:size}\includegraphics[width=0.5\linewidth]{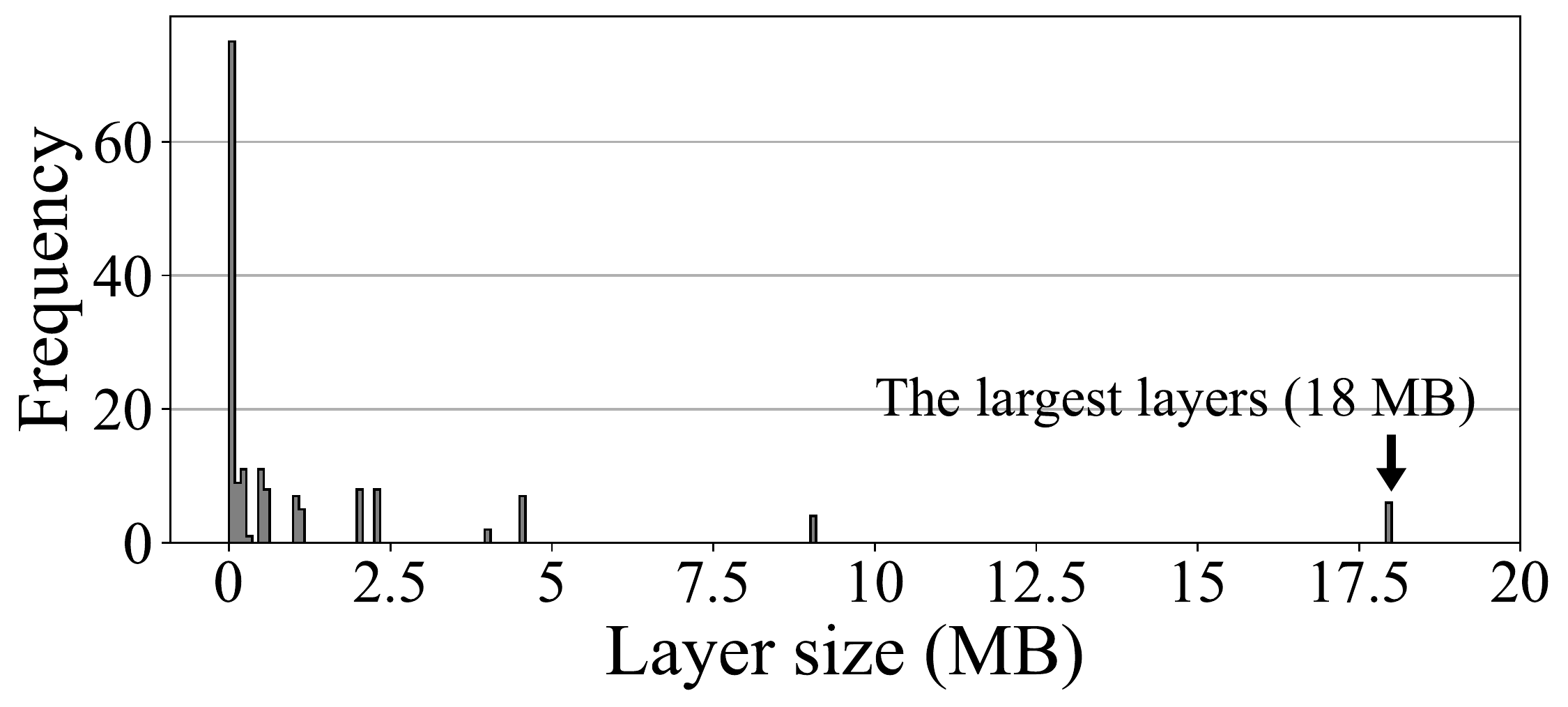}}
\caption{Measured distributions.}
\label{fig:hist}
\vspace{-0.2cm}
\end{figure}

In YOLOv4 in Fig.~\ref{fig:layer}, we have $N$ = 162 layers with various layer types (i.e., 110 convolutional, 21 route, 23 shortcut, 3 maxpool, 3 yolo, and 2 upsample layers). The total size of the layers is 245.8~MB. As shown in the figure, Most $r_i$, $c_i$, and $k_i$ have strong correlations to $s_i$, because large layers naturally involve more time to read, copy, and execute. One interesting observation is that the foremost layers exhibit relatively large kernel times, which do not properly reflect their small layer sizes. The reason is that the foremost layers extract features from the image, so their kernel times tend to depend strongly on the input image size rather than the layer size.

\begin{wrapfigure}{r}{0.25\textwidth}
\vspace{-0.4cm}
  \centering{\includegraphics[width=0.25\textwidth]{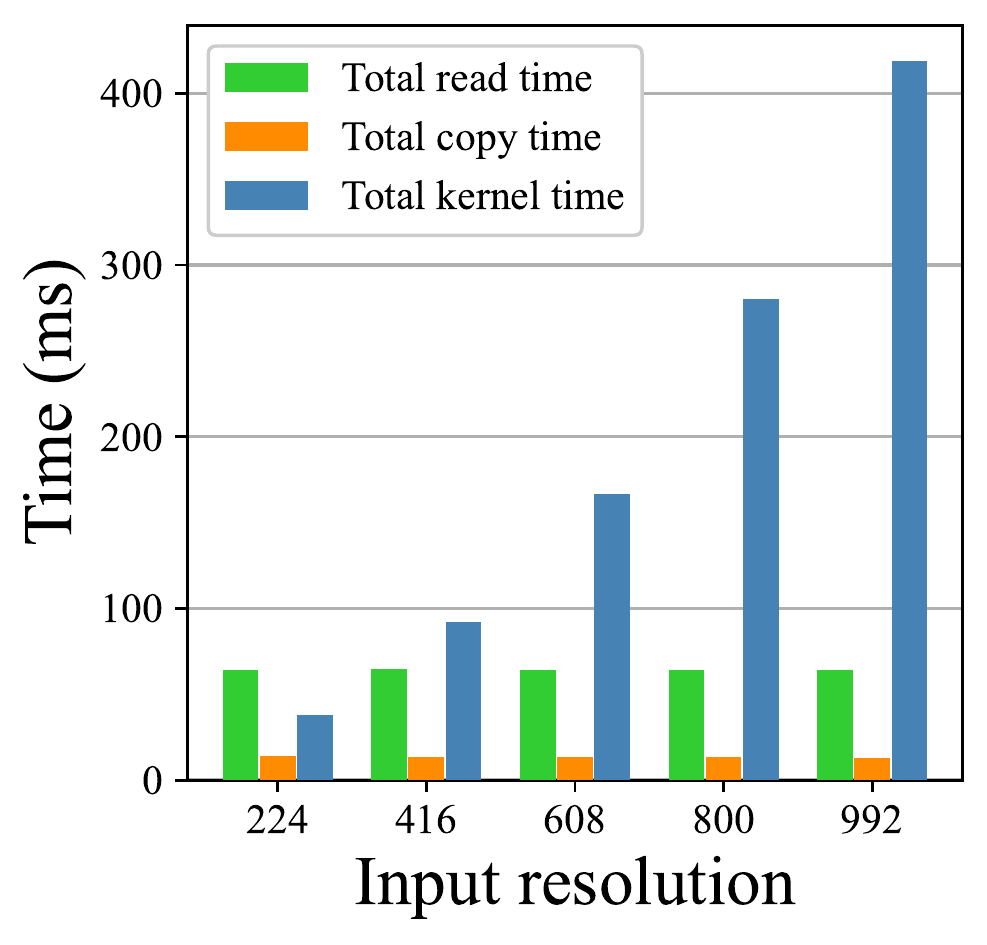}}
  \caption{How the input resolution affects the timings (YOLOv4).}
  \label{fig:resolution}
\end{wrapfigure}

By looking at Fig.~\ref{fig:hist}, all the four distributions are long-tailed.
In particular, regarding the layer size, 75 layers (46\%) are under 0.1~MB, while we have six layers with the same largest layer size $s^{max}$ = 18.0~MB.
Another interesting question is how the input image resolution affects the timings. Fig.~\ref{fig:resolution} compares the total read, copy, and kernel times with varying input resolutions. For example, 608 in the x-axis represents a 608 $\times$ 608 input resolution to the DNN. The figure indicates that the input resolution has no noticeable impact on the read and copy times because they only depend on the layer size. However, the kernel time is heavily dependent on the input resolution. Thus, if the input resolution gets too small (e.g., the 224 $\times$ 224 case), the total kernel time becomes even smaller than the total read time, possibly losing the optimization opportunity for overlapping read and copy operations behind kernel executions. However, such low input resolutions are practically not used due to their inferior detection accuracy. We use 608 $\times$ 608 as the default input resolution, which is from the YOLO DNN's default configuration.


\subsection{Problem Description} \label{sec:motivation}

In the preloading architecture, all the read and copy operations are finished in the initialization phase. Thus, only the kernel operations are executed in the inference phase. Thus, its delay is given by
\begin{equation}
\sum_{i = 1}^{N}{k_i},
\label{eq:d_preloading}
\end{equation}
which is the {\em optimal delay} that cannot be reduced any further. In the memory perspective, we need a CPU buffer and a GPU buffer, both of which should be able to accommodate the entire model parameters. Thus, the memory consumption is two times the total layer size, given by
\begin{equation}
2 \times \sum_{i = 1}^{N}{s_i}.
    \label{eq:m_preload}
\end{equation}

Beginning from the above preloading architecture, this study has the following objectives:
\begin{itemize}
    \item To design and implement a layer-by-layer loading and execution (Demand Layering) framework (Section~\ref{sec:demand}).
    \item To minimize the delay overhead caused by the additional read and copy operations (Section~\ref{sec:pipeline} and Section~\ref{sec:bleeding}).
    \item To evaluate our implementation in terms of memory and delay with state-of-the-art DNN workloads (Section~\ref{sec:experiments}).
\end{itemize}
This study limits our problem to systems running a single DNN in isolation. However, once solved, the solution can be easily applied to systems concurrently running multiple DNNs. In fact, our solution is more valuable in such multi-DNN systems that require significantly larger memory to encompass multiple DNNs~\cite{bai2020pipeswitch}.

\section{Demand Layering} \label{sec:demand}

\begin{figure}
\centering
    \subfloat[\centering{Cached I/O.}]{\label{fig:paged}
    \includegraphics[width=0.11\textwidth]{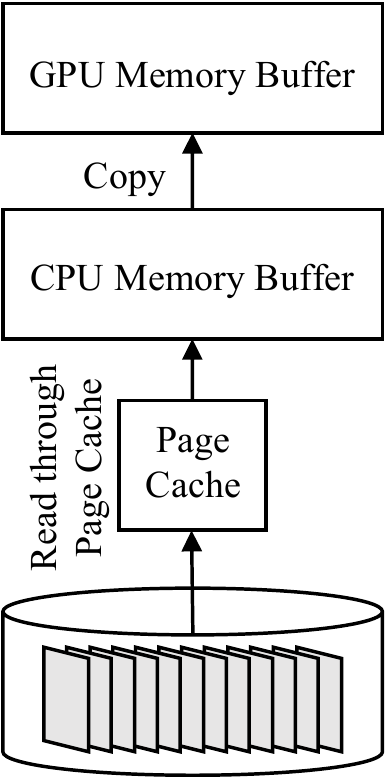}}\hspace{0.4cm}
    \subfloat[Direct I/O.]{\label{fig:direct}
    \includegraphics[width=0.11\textwidth]{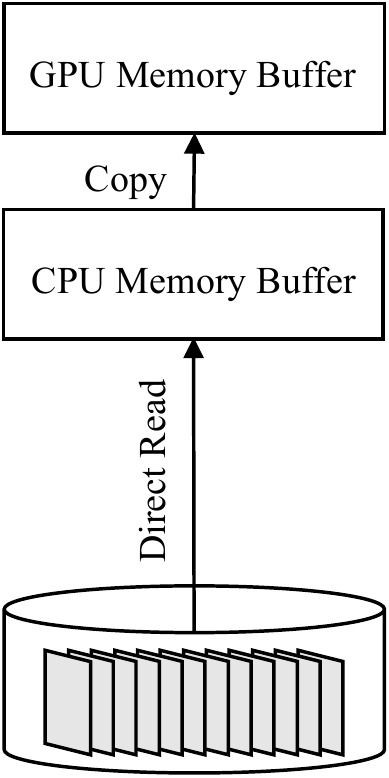}}\hspace{0.4cm}
    \subfloat[To zero-copy memory.]{\label{fig:zero}
    \includegraphics[width=0.11\textwidth]{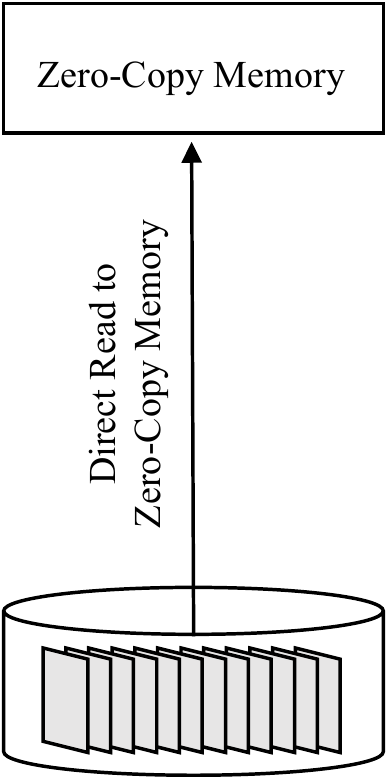}}
\caption{Comparison of I/O methods for read operations.}
\label{fig:io_methods}
\end{figure}

\subsection{Read Operations}

\begin{figure}
    \centering 
    \subfloat[Read distributions.]
    {\label{fig:read_hist}\includegraphics[width=0.33\linewidth]{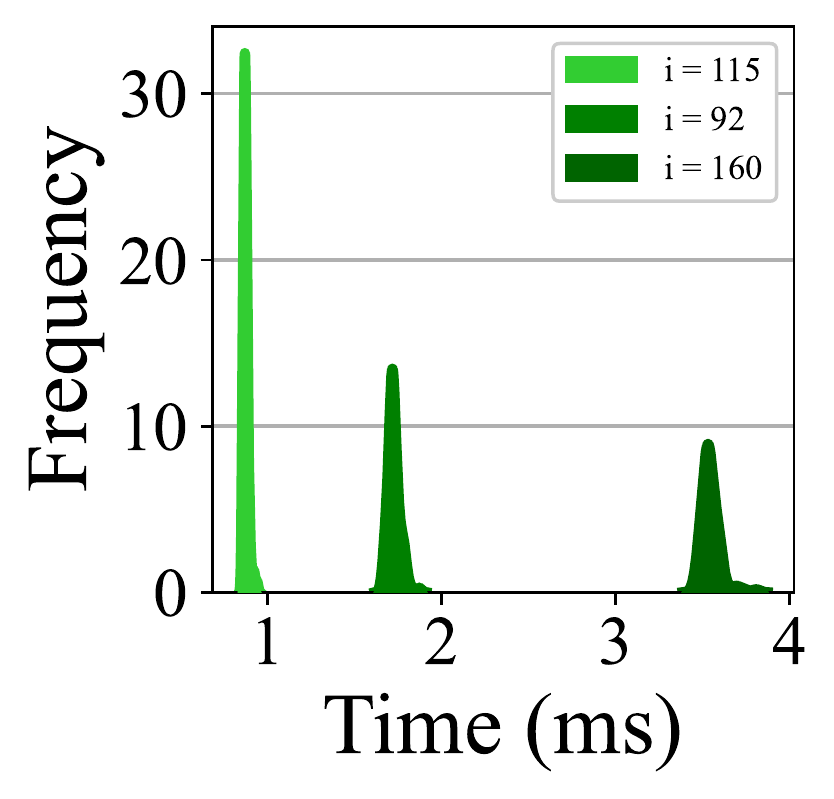}}
    \subfloat[Copy distributions.]
    {\label{fig:copy_hist}\includegraphics[width=0.33\linewidth]{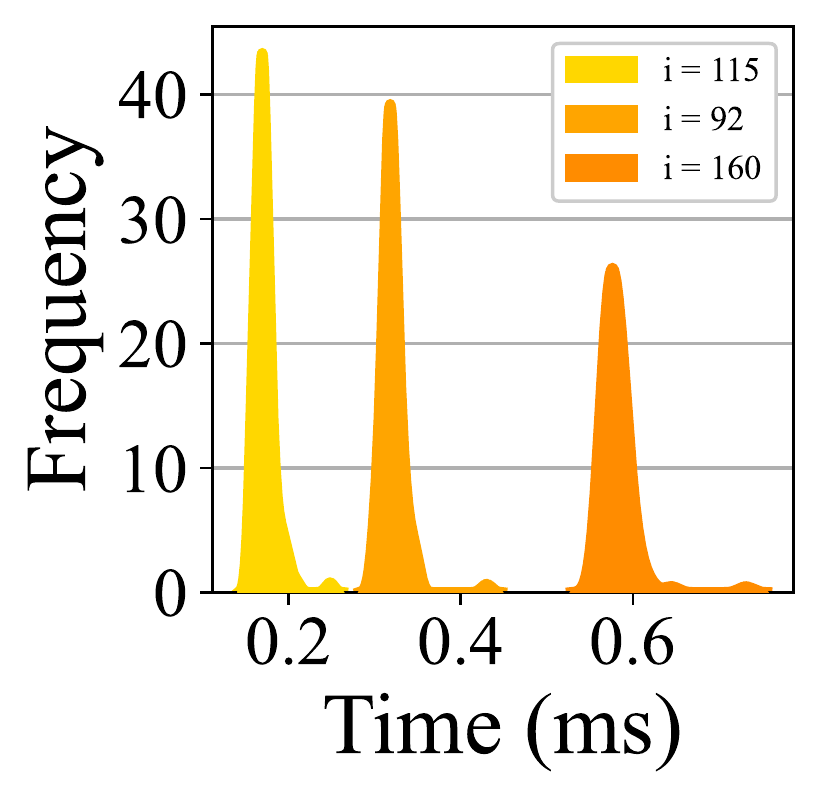}}
    \subfloat[Kernel distributions.]
    {\label{fig:kernel_hist}\includegraphics[width=0.33\linewidth]{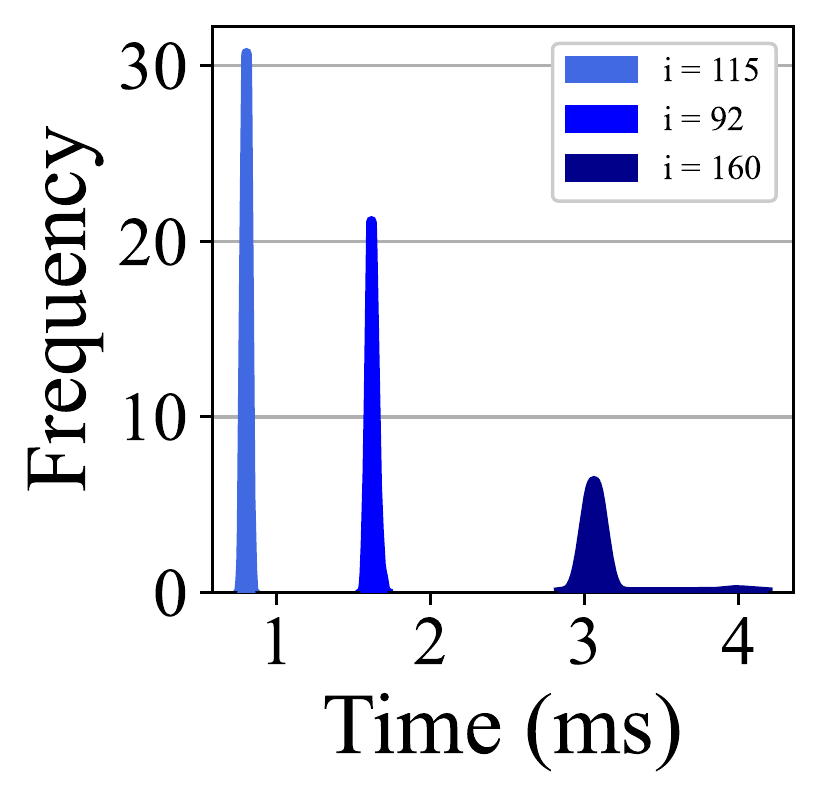}}\\
\caption{Profiling results of selected convolutional layers ($s_{115}$=4~MB, $s_{92}$=9~MB, and $s_{160}$=18~MB).}
\label{fig:profile}
\end{figure}

A read operation $R_i$ is to read the $l_i$'s portion (layer parameters) of the model file to CPU memory. We assume that the model file is sequential such that there need no random offset changes between reads to fully exploit the SSD's sequential read performance. For the sake of real-time performance, the variance of $r_i$ must be kept within a predictable range. However, assuming the stock Linux scheduler, it is not possible to provide hard guarantees.

Nonetheless, to minimize the variance as much as possible, our choice is to use {\em direct I/O} (Fig.~\ref{fig:direct}) instead of {\em cached I/O} (Fig.~\ref{fig:paged}), expecting to suppress unpredictable cache behaviors. Fig.~\ref{fig:read_hist} shows the profiling results of read times by direct I/O with selected YOLOv4 layers, showing that the variances are predictable within ranges. Also, by using direct I/O, we have another significant benefit of saving CPU memory for the page cache. Due to the aggressive page cache policy of Linux, the page cache will eventually duplicate most contents of the model file in CPU memory. Because we intend to minimize the memory usage by model parameters, it is desirable not to have a duplicate in the page cache.


In addition, our read operations are designed as asynchronous, using the POSIX asynchronous I/O method (AIO). With AIO, the calling thread does not block while a background read thread is processing pending read requests in the AIO queue. Since AIO has a limitation of supporting only \texttt{O\_DIRECT} files, there is no choice but to use direct I/O.

\subsection{Copy Operations} \label{sec:copy}

A copy operation $C_i$ is to transfer a layer's parameters in CPU memory to GPU memory, making them accessible by GPU kernels. The CUDA runtime provides synchronous (\texttt{cudaMemcpy()}) and asynchronous (\texttt{cudaMemcpyAsync()}) copy functions. Here, our choice is, again, to make our copy operations fully asynchronous such that the completion of an asynchronous copy request can be later checked by CUDA synchronization primitives like the CUDA event API.

However, the asynchronous copy functions may sometimes block depending on the CPU memory buffer types. The CUDA runtime provides two basic memory types for CPU memory buffers: (i) {\em pageable} and (ii) {\em host-pinned}. A pageable buffer is allocated by \texttt{malloc()}, while a host-pinned buffer is allocated by \texttt{cudaHostAlloc()}. Invoking an asynchronous copy with its source location at a pageable memory buffer is unsafe. The copy will fail if the source buffer is paged out during the copy operation. In this regard, the CUDA runtime provides a hidden staging area, as shortly explained with Fig.~\ref{fig:flow}, which is in host-pinned memory that is never paged out. Although the internal organization of the staging area is not officially released, we found out that, by reverse engineering, it is made of a limited number of fixed-length buffers. If the buffers are all in use, the copy request will block. Also, if a copy size is larger than the largest individual staging buffer, it will block. With experiments, we observed many such unexpected blocking scenarios when copying large layers with asynchronous copy functions. With the above observation, our choice is to use host-pinned memory for CPU buffer allocations. Fig.~\ref{fig:copy_hist} shows the profiling results of such copy times of selected YOLOv4 layers, which are much smaller than read and kernel times.


\subsection{Kernel Operations}

A kernel operation $K_i$ is to execute CUDA kernel functions that implement the corresponding layer, accessing only GPU memory buffers allocated by \texttt{cudaMalloc()}. For example, if the layer is a convolutional layer, it may call CUDA-based GeMM (General Matrix Multiplication) functions with input feature maps and layer parameters (i.e., convolution filters) to produce output feature maps. If a higher-level DNN library (e.g., cuDNN) is used, it may call the library's built-in implementation of a convolutional layer. Fig.~\ref{fig:kernel_hist} shows the profiling results of kernel times of selected YOLOv4 layers, which are significantly larger than the read and copy times. Kernel requests are non-blocking in its nature. A kernel request just places the request in a CUDA stream (i.e., a queue for pending kernel calls) and returns. Then the kernel request eventually goes through the queue to be executed by the GPU execution engine. To check the completion of a kernel operation, we put a CUDA event, a synchronization primitive, behind each kernel request in the same CUDA stream. Later, the completion of each kernel operation can be confirmed by querying the inserted CUDA event.

\subsection{Scheduling Read, Copy, and Kernel Operations} \label{sec:scheduling}

\begin{figure}
    \centering
    \includegraphics[width=0.4\textwidth]{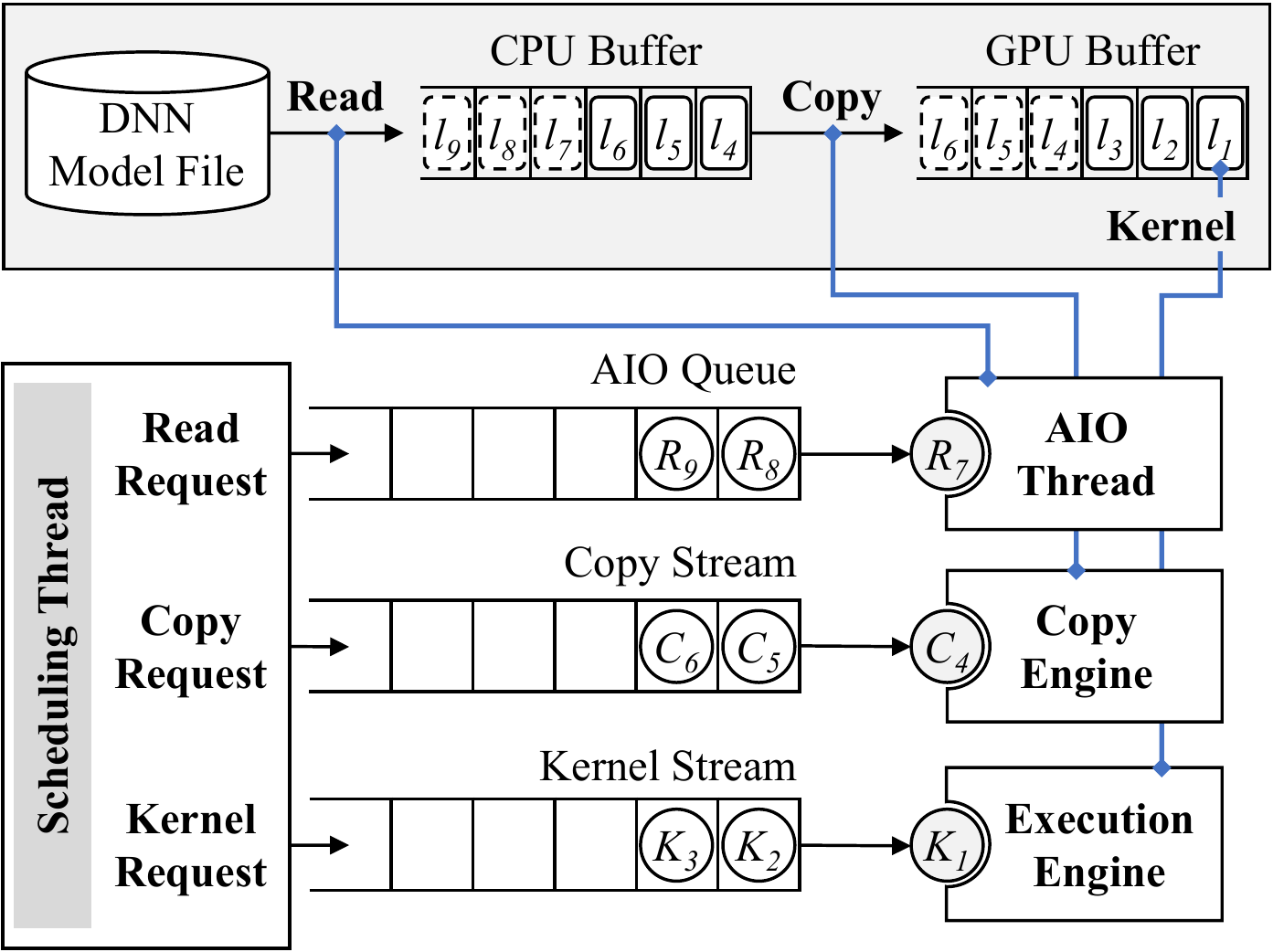}
    \caption{Scheduling of read, copy, and kernel operations.}
    \label{fig:scheduling}
\end{figure}

Fig.~\ref{fig:scheduling} shows the scheduling architecture of Demand Layering, where a scheduling thread (on CPU) {\em releases} asynchronous read, copy, and kernel requests. Once requested, the pending read requests in the AIO queue are sequentially processed by the AIO thread; the pending copy requests in the {\em copy stream} are processed by the GPU copy engine; the pending kernel requests in the {\em kernel stream} are executed by the GPU execution engine. Since all the read, copy, and kernel requests are asynchronous, the scheduling thread can freely release a request at any time without any blocking.

Even with the freedom in scheduling, there are strict scheduling constraints to maintain the system's reliability. We define three kinds of such constraints: (i) {\em release constraints}, (ii) {\em completion constraints}, and (iii) {\em target buffer constraints}. A release constraint is denoted by $\mathcal{R}_{prev} \rightarrow \mathcal{R}_{next}$, where an arbitrary request $\mathcal{R}_{next}$ cannot be released before the release of $\mathcal{R}_{prev}$. A completion constraint is denoted by $\mathcal{R}_{prev} \mapsto \mathcal{R}_{next}$, where a request $\mathcal{R}_{next}$ cannot be released before the completion of $\mathcal{R}_{prev}$. A target buffer constraint is denoted by $\mathcal{R} \leadsto CPU(s)$ or $\mathcal{R} \leadsto GPU(s)$, where a request $\mathcal{R}$ can be released only when a contiguous memory area with the size of $s$ in the target CPU memory or GPU memory is available. Then our scheduling constraints can be formally defined as follows:
\begin{equation}
\begin{split}
    &R_i \rightarrow R_{i+1}, C_i \rightarrow C_{i+1}, K_i \rightarrow K_{i+1},\\
    &R_i \mapsto C_{i}, C_i \mapsto K_{i},\\
    &R_i \leadsto CPU(s_i), C_i \leadsto GPU(s_i),
\end{split}
\end{equation}
meaning that every release of a request in the same type must be in the strict order of layers; a copy request can only be released after that layer's read operation is completed; a kernel request can only be released after its corresponding copy operation is completed; a read request cannot be released until the available CPU memory is enough for the layer size; a copy request cannot be released until the available GPU memory is enough for the layer size.



\section{Pipeline Schedule Optimization} \label{sec:pipeline}

\begin{figure*}
    \centering
    \includegraphics[width=0.9\textwidth]{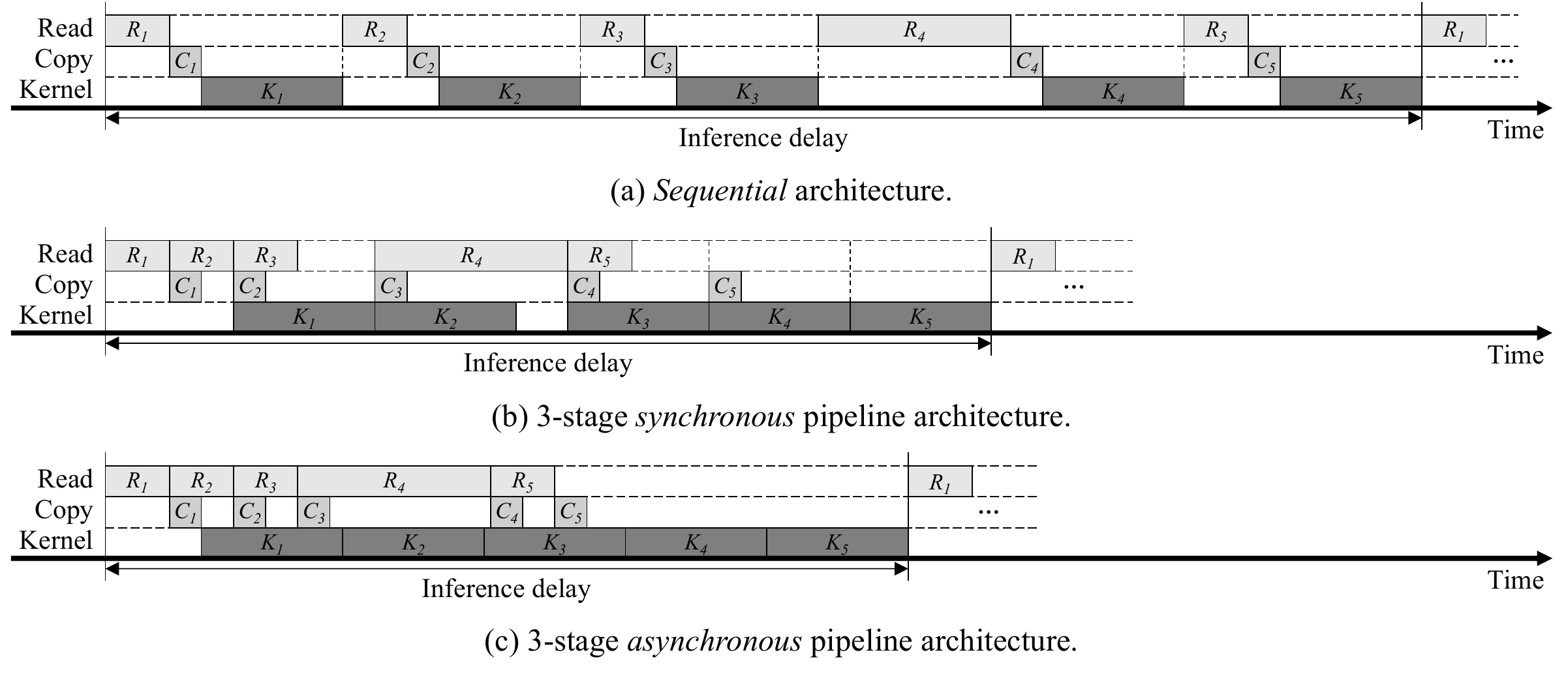}
    \caption{Pipeline scheduling algorithms, illustrated by a simplified 5-layer DNN.}
    \label{fig:pipeline_design}
\end{figure*}

\subsection{Sequential Architecture} \label{sec:sequential}

The most na\"ive scheduling method for Demand Layering is to execute read, copy, and kernel operations sequentially without overlapping their executions, as in Fig~\ref{fig:pipeline_design}a. In this method, we just need a CPU buffer and a GPU buffer both with the size of $s^{max}$, making its memory requirement fall from the preloading architecture's $2 \times \sum_{i = 1}^{N}{s_i}$ in Eq.(\ref{eq:m_preload}) to 
\begin{equation}
2 \times s^{max},
\end{equation}
which is a significant drop from the order of the entire model to the order of a layer. For example, YOLOv4 has a 92.7\% reduction (from 491.6~MB to 36.0~MB). However, the sequential architecture causes a significant amount of delay overhead, as illustrated in Fig~\ref{fig:pipeline_design}a by the GPU idling intervals during read and copy operations, increasing the inference delay to
\begin{equation}
\sum_{i = 1}^{N}{(r_i + c_i + k_i)}
\end{equation}
from the preloading architecture's $\sum_{i = 1}^{N}{k_i}$ in Eq.(\ref{eq:d_preloading}).

\subsection{Synchronous Pipeline} \label{sec:syncpipe}

To minimize the delay overhead of the sequential architecture, our next approach is to employ a pipeline architecture as illustrated in Fig.~\ref{fig:pipeline_design}b. This pipeline architecture is possible due to the parallelizable nature of read, copy, and kernel executions. In this 3-stage {\em synchronous} pipeline architecture, the three pipeline stages (i.e., read, copy, and kernel stages) advance with a single synchronized pipeline cycle. To implement such architecture, we need to employ the {\em double-buffering} scheme because $R_i$ must be writing to a CPU buffer while $C_{i-1}$ is moving data from the same CPU buffer to a GPU buffer simultaneously. The same applies to GPU memory because $C_{i-1}$ must be writing to a GPU buffer while $K_{i-2}$ is executing its layer in the same GPU buffer. Due to this double-buffering constraint, its memory requirement is increased to
\begin{equation}
2 \times 2 \times s^{max},
\end{equation}
which is double the sequential architecture's requirement.

Despite the increased memory requirement, due to the pipeline architecture, most read and copy operations are hidden behind kernel operations since kernel operations are mostly the longest. However, certain long read operations, such as $R_4$ in Fig.~\ref{fig:pipeline_design}b, can dominate certain pipeline cycles. Considering such scenarios, the inference delay can be generally calculated as
\begin{equation}
\begin{split}
r_1 + \textrm{max}(\{r_2, c_1\}) &+ \sum_{i=1}^{N-2}{\textrm{max}(\{r_{i+2}, c_{i+1}, k_{i}\})} \\
&+ \textrm{max}(\{c_N, k_{N-1}\}) + k_{N},
\end{split}
\end{equation}
by incorporating all the pipeline cycles ranging from $R_1$ to $K_N$. As illustrated in Fig.~\ref{fig:pipeline_design}b, this pipelined execution significantly reduces the delay overhead from the sequential architecture.


\subsection{Asynchronous Pipeline} \label{sec:asyncpipe}

To minimize the possible GPU idling intervals in the synchronous pipeline architecture, our next approach is to employ the asynchronous pipeline~\cite{nowick2011high}, where pipeline stages advance at their own paces without being synchronized by a single pipeline cycle. Fig~\ref{fig:pipeline_design}c shows how the asynchronous pipeline differs from the synchronous one in Fig~\ref{fig:pipeline_design}b. As illustrated in the figure, $R_4$ can be released right after the completion of $R_3$, which in turn eliminates the GPU idling interval between $K_2$ and $K_3$ that existed in the synchronous architecture. In the asynchronous pipeline architecture, we can reduce the memory requirement to
\begin{equation}
2 \times s^{max},
\end{equation}
since we no longer need the double-buffering scheme. Only a single pair of CPU and GPU buffers, both with the size of $s^{max}$, suffices to execute the entire DNN. In contrast to the buffers in the synchronous architecture, which store only one layer at a time, the buffers in the asynchronous architecture are designed as {\em circular queues} that can hold multiple pending layers, as illustrated in the top of Fig.~\ref{fig:scheduling}.

\begin{figure*}
    \centering
    \includegraphics[width=0.9\textwidth]{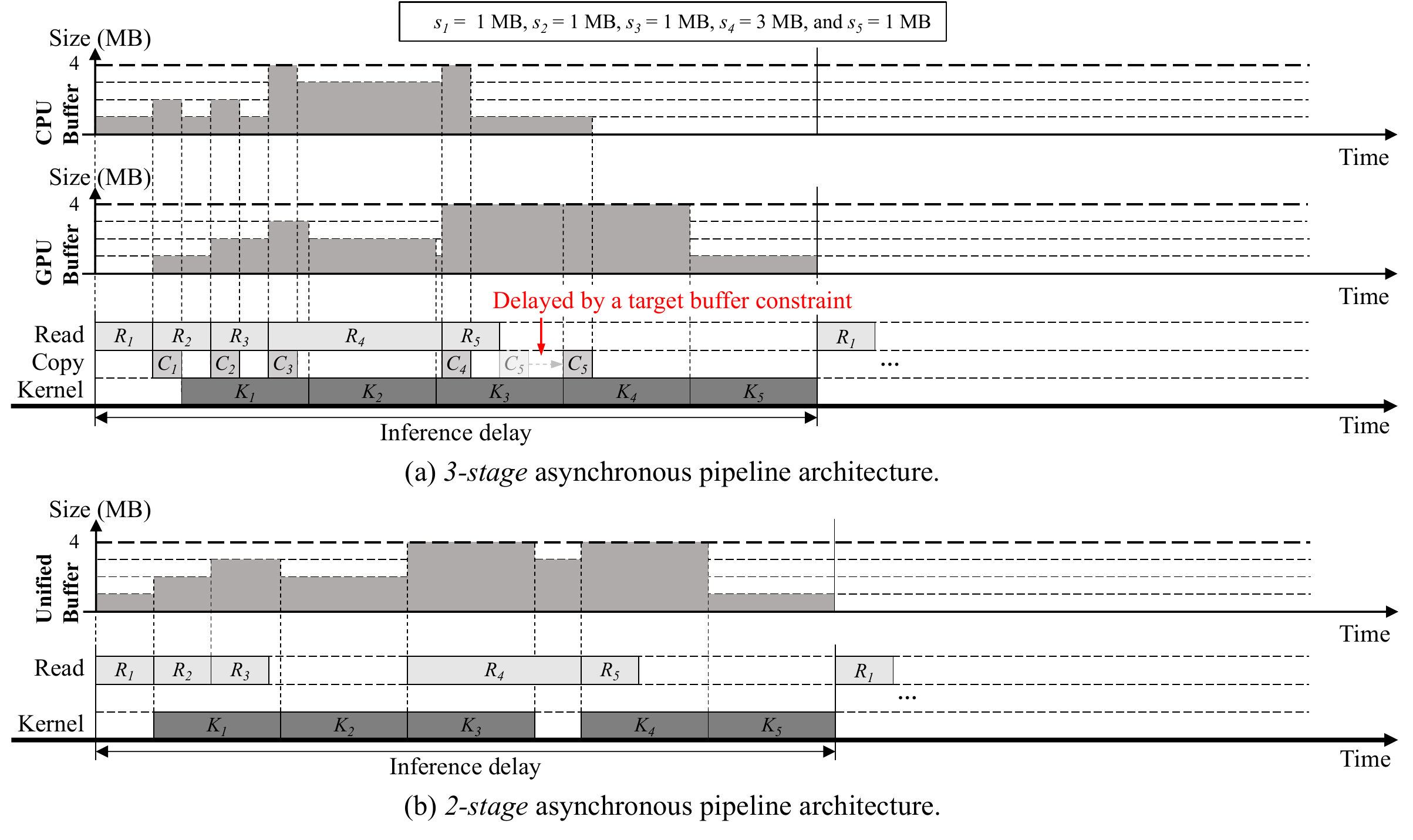}
    \caption{3-stage vs. 2-stage pipeline architecture.}
    \label{fig:2stagepipe}
\end{figure*}

\begin{algorithm}
\begin{algorithmic}[1]
\Require $\mathbb{R}=\{R_1, R_2, \cdots, R_N\}$
\Require $\mathbb{C}=\{C_1, C_2, \cdots, C_N\}$
\Require $\mathbb{K}=\{K_1, K_2, \cdots, K_N\}$
\Require $b^{c}$ \ \ \ \ \ $\triangleright$ Free CPU buffer (queue) size
\Require $b^{g}$ \ \ \ \ \ $\triangleright$ Free GPU buffer (queue) size
\State $h, i, j, k, \leftarrow 1$
\While {$\mathbb{R} \neq \emptyset$ {\bf or} $\mathbb{C} \neq \emptyset$ {\bf or} $\mathbb{K} \neq \emptyset$}

    \If {\Call {IsFinished}{$K_h$}}
        \State $b^g \leftarrow b^g + s_h$
        \State $h \leftarrow h + 1$
         \ \ \ \ \ $\triangleright$ $h$: foremost kernel index
    \EndIf

    \If {$b^{c} \geq s_i$}\\
        \ \ \ \ \ \ \ \ \Call {RequestAsync}{$R_i$}
        \State $\mathbb{R} \leftarrow \mathbb{R} \setminus \{R_i\}$
        \State $b^c \leftarrow b^c - s_i$
        \State $i \leftarrow i + 1$
        \ \ \ \ \ $\triangleright$ $i$: next read index to be requested
    \EndIf

    \If {\Call {IsFinished}{$R_j$} {\bf and} $b^g \geq s_j$}
        \State \Call {RequestAsync}{$C_j$}
        \State $\mathbb{C} \leftarrow \mathbb{C} \setminus \{C_j\}$
        \State $b^g \leftarrow b^g - s_j$
        \State $j \leftarrow j + 1$
                \ \ \ \ \ $\triangleright$ $j$: next copy index to be requested
    \EndIf
    
    \If {\Call {IsFinished}{$C_k$}}\\
        \ \ \ \ \ \ \ \ \Call {RequestAsync}{$K_k$}
        \State $\mathbb{K} \leftarrow \mathbb{K} \setminus \{K_k\}$
        \State $b^c \leftarrow b^c + s_k$
        \State $k \leftarrow k + 1$
                        \ $\triangleright$ $k$: next kernel index to be requested
    \EndIf
\EndWhile
\end{algorithmic}
\caption{Asynchronous pipeline scheduling}
\label{alg:async}
\end{algorithm}

However, due to its complexity, the scheduling thread for the asynchronous pipeline architecture should be carefully designed considering all the scheduling constraints in Section~\ref{sec:scheduling}. Algorithm~\ref{alg:async} describes our scheduling algorithm. The while loop is a busy loop deciding when to release requests asynchronously until there remains no more operation to schedule. The algorithm handles four indices, where $i$, $j$, and $k$ denote the read, copy, and kernel requests next to be scheduled. In addition, $K_h$ denotes the foremost kernel operation that is executing in GPU. {\bf Lines 3-6} take care of the completion of the currently executing kernel operation by freeing the allocated amount of GPU buffer and advancing the index $h$. The constraint $R_i \leadsto CPU(s_i)$ is checked in {\bf Lines 7-12} for submitting a read request. {\bf Lines 13-18} check the constraints $R_j \mapsto C_{j}$ and $C_j \leadsto GPU(s_j)$ for a copy request. {\bf Lines 19-24} also consider the constraint $C_k \mapsto K_{k}$ for a kernel request.

\subsection{Iterative Optimization by Memory-Delay Tradeoff} \label{sec:tradeoff}

In the asynchronous pipeline architecture, we have the freedom to increase buffer sizes beyond the minimum requirement to exploit the memory-delay tradeoff. For example, if the buffer sizes are large enough such that the entire model can be brought into memory, the inference delay can be very close to the optimal delay of the preloading architecture. With this insight, we additionally propose an iterative delay minimization method. In this method, the optimization process begins from the asynchronous pipeline's minimal memory requirement ($2 \times s^{max}$) and iteratively increases the buffer sizes with a step size (denoted by $\delta$) until the delay does not decrease any more. We call this a {\em minimal delay} configuration.

To be precise, the iterative optimization method has two phases for finding the minimal delay point. The first phase records every resulting average delay by gradually increasing the buffer sizes until the maximum (i.e., the model size), by which the minimal delay point can be found. In some cases, multiple points can have the same minimal delay since the delay is not monotonically decreasing. Then, in the second phase, the recorded results are revisited from the beginning, which stops when we first encounter the minimal delay.

\section{Bleeding-Edge Optimization for Xavier SoCs} \label{sec:bleeding}

{\bf Note:} This section further optimizes the memory usage, assuming some features of the most brand new Nvidia Xavier SoCs. Thus, this section's optimization method can possibly cause unknown stability issues in other hardware platforms.

{\bf Zero-copy memory.} The CUDA runtime provides {\em unified memory} that provides a single address space across CPU and GPU memory, eliminating the need for copies in the programmer's perspective. However, in dGPU systems, copy operations are unavoidable. Thus, they are executed in the background by the CUDA runtime. In contrast, in iGPU systems, {\em zero-copy} memory management can be implemented since CPU and GPU share a common memory device, as in Fig.~\ref{fig:zero}. Our experimental platform provides the following zero-copy memory types: (i) host-pinned zero-copy memory by \texttt{cudaHostAlloc()} with a \texttt{cudaHostAllocMapped} option and (ii) managed memory by \texttt{cudaMallocManaged()} with a \texttt{cudaMemAttachHost} option. They are almost identical, except that managed memory reportedly does not allow simultaneous accesses from CPU and GPU~\cite{bateni2020co}. However, our experiment found that it works just as we wanted, even with slightly better performance than host-pinned zero-copy memory. Thus, we decided to use managed memory.

{\bf Two-stage pipeline.} With zero-copy memory, we no longer need the copy stage, enabling a 2-stage pipeline. Fig.~\ref{fig:2stagepipe} compares the difference between the 3-stage and 2-stage asynchronous pipelines. The figure also illustrates how the buffers are utilized. Note that the schedule in Fig.~\ref{fig:2stagepipe}a is slightly different from Fig.~\ref{fig:pipeline_design}c, because Fig.~\ref{fig:2stagepipe}a additionally considers the target buffer constraints ($R_i \leadsto CPU(s_i)$ and $C_i \leadsto GPU(s_i))$, assuming 4~MB buffer sizes, which were not considered in Fig.~\ref{fig:pipeline_design}c. More specifically, in Fig.~\ref{fig:2stagepipe}a, $C_5$ cannot be released right after $R_5$, because between the completion of $R_5$ and the completion of $K_3$, the GPU buffer is full. After the completion of $K_3$, $C_5$ can be released after reserving the just freed 1~MB space. In this method, the memory requirement is reduced to
\begin{equation}
s^{max},
\end{equation}
which is {\em memory-optimal} with just a single layer size. There might be a misunderstanding that the 2-stage architecture will provide shorter delays than the 3-stage counterpart. By comparing Fig.~\ref{fig:2stagepipe}a and Fig.~\ref{fig:2stagepipe}b, however, the delay of the 2-stage architecture is even larger than the 3-stage architecture, which is due to the reduced total buffer size that makes the read operations feel more challenging to run due to the deteriorated target buffer constraint. For example, $R_4$ is heavily delayed by its target buffer constraint, leading to a GPU idling interval between $K_3$ and $K_4$ as a consequence. Thus, the primary objective of the 2-stage pipeline is to make the {\em minimal memory} configuration, not to reduce the inference delay.

\begin{figure*}
\centering
    \subfloat[\centering{YOLOv3.}]{\label{fig:v3_archi}
    \includegraphics[width=0.19\textwidth]{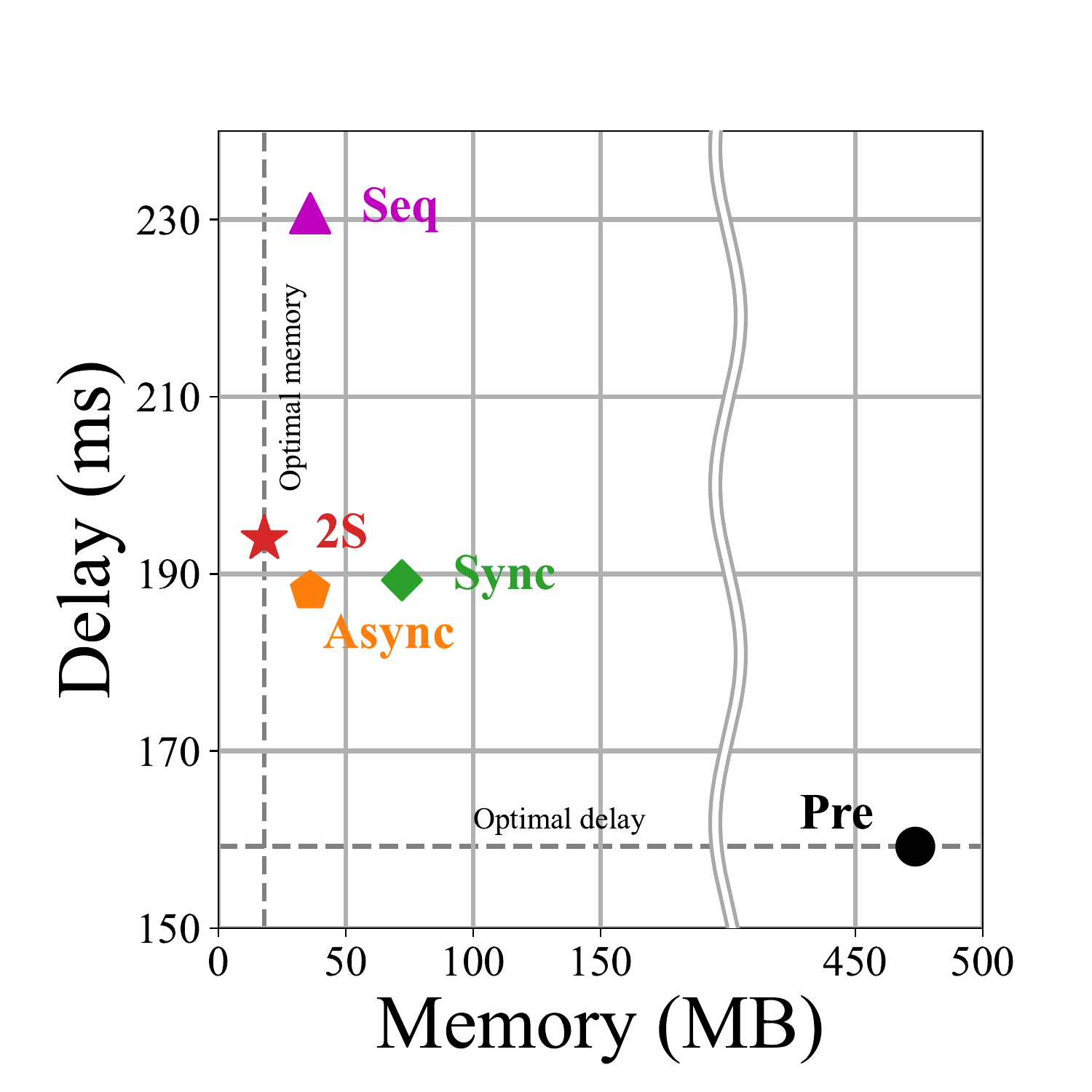}}\hspace{-0.1cm}
    \subfloat[YOLOv4.]{\label{fig:v4_archi}
    \includegraphics[width=0.19\textwidth]{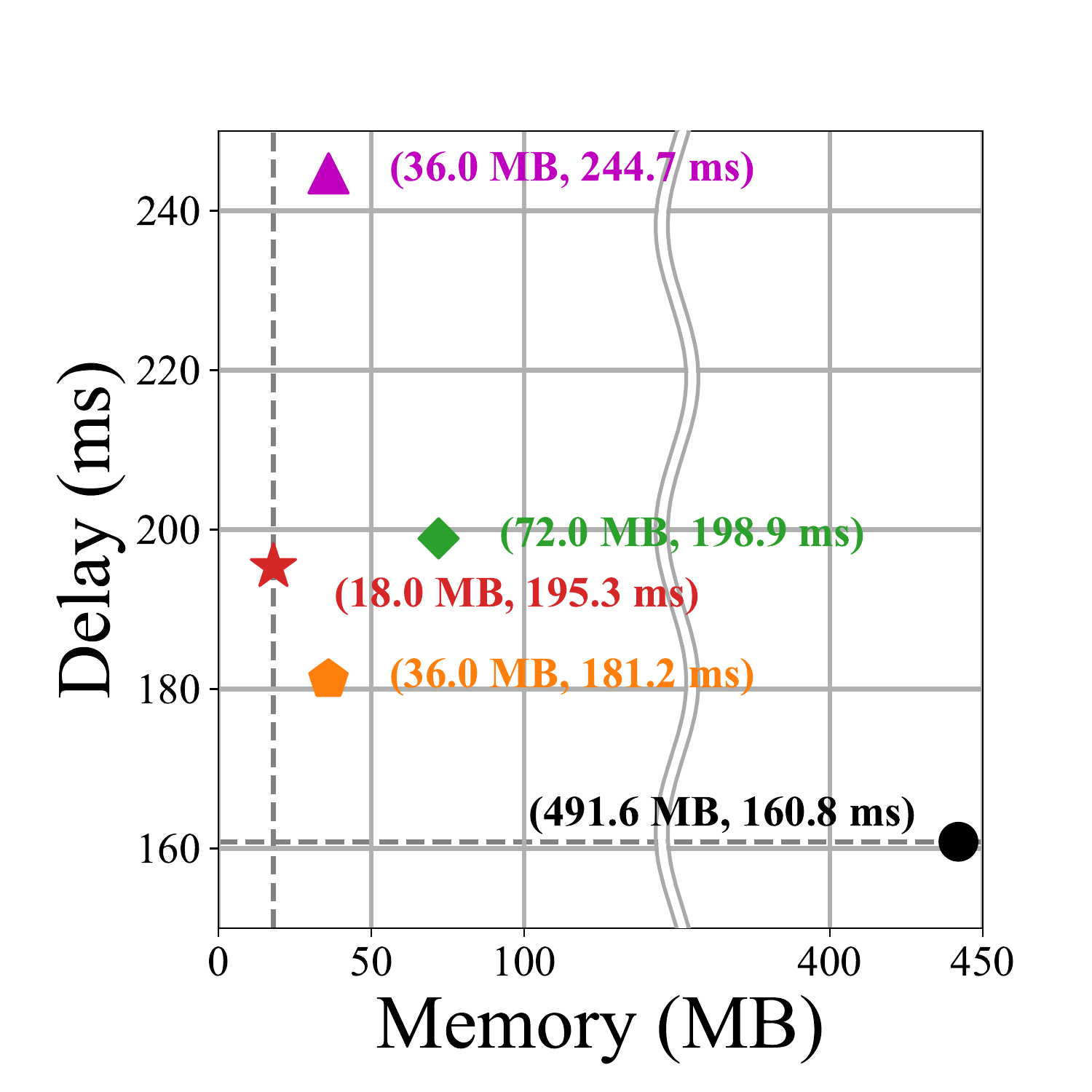}}\hspace{-0.1cm}
    \subfloat[YOLOv4-P6.]{\label{fig:v4p6_archi}
    \includegraphics[width=0.19\textwidth]{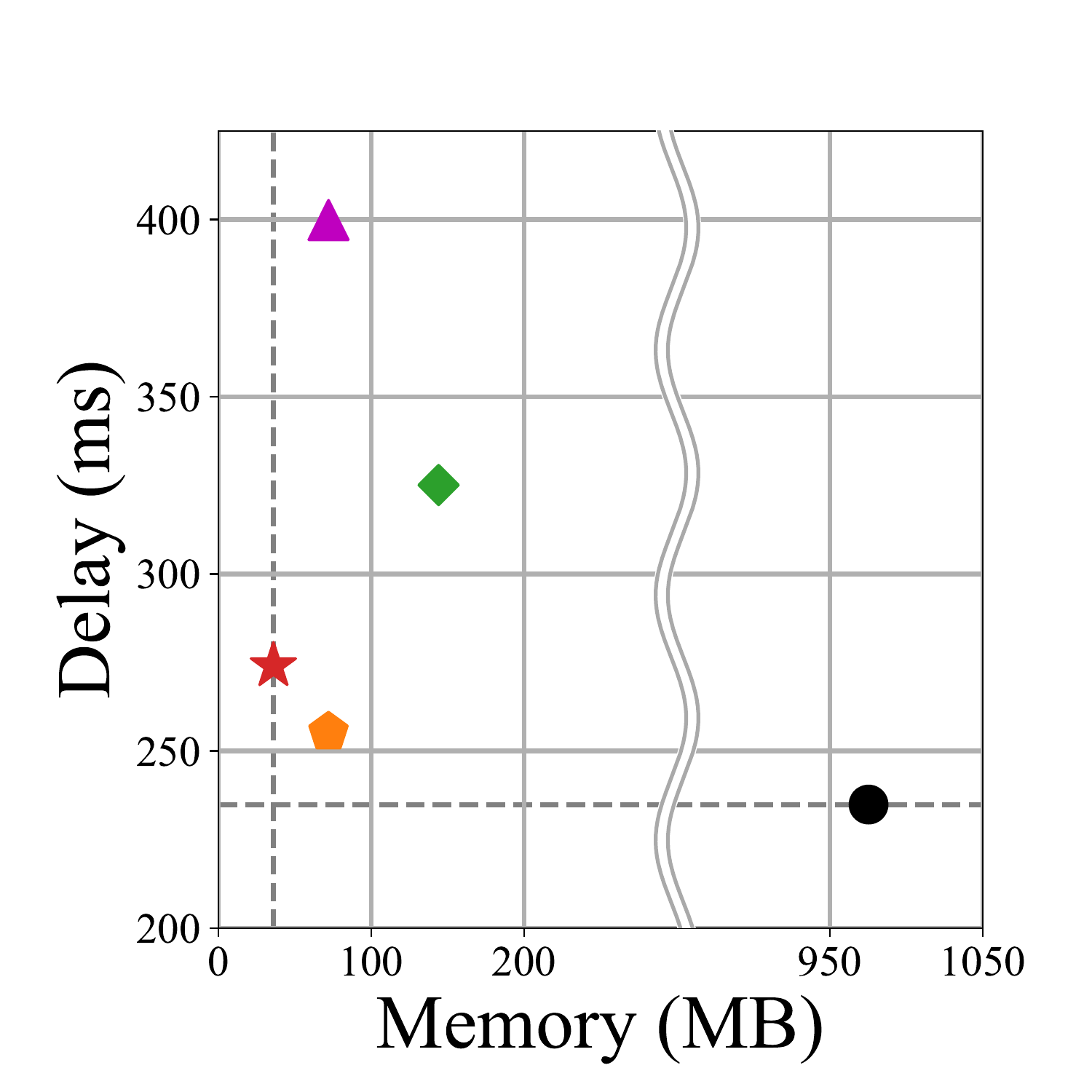}}\hspace{-0.1cm}
    \subfloat[ResNet.]{\label{fig:resnet_archi}
    \includegraphics[width=0.19\textwidth]{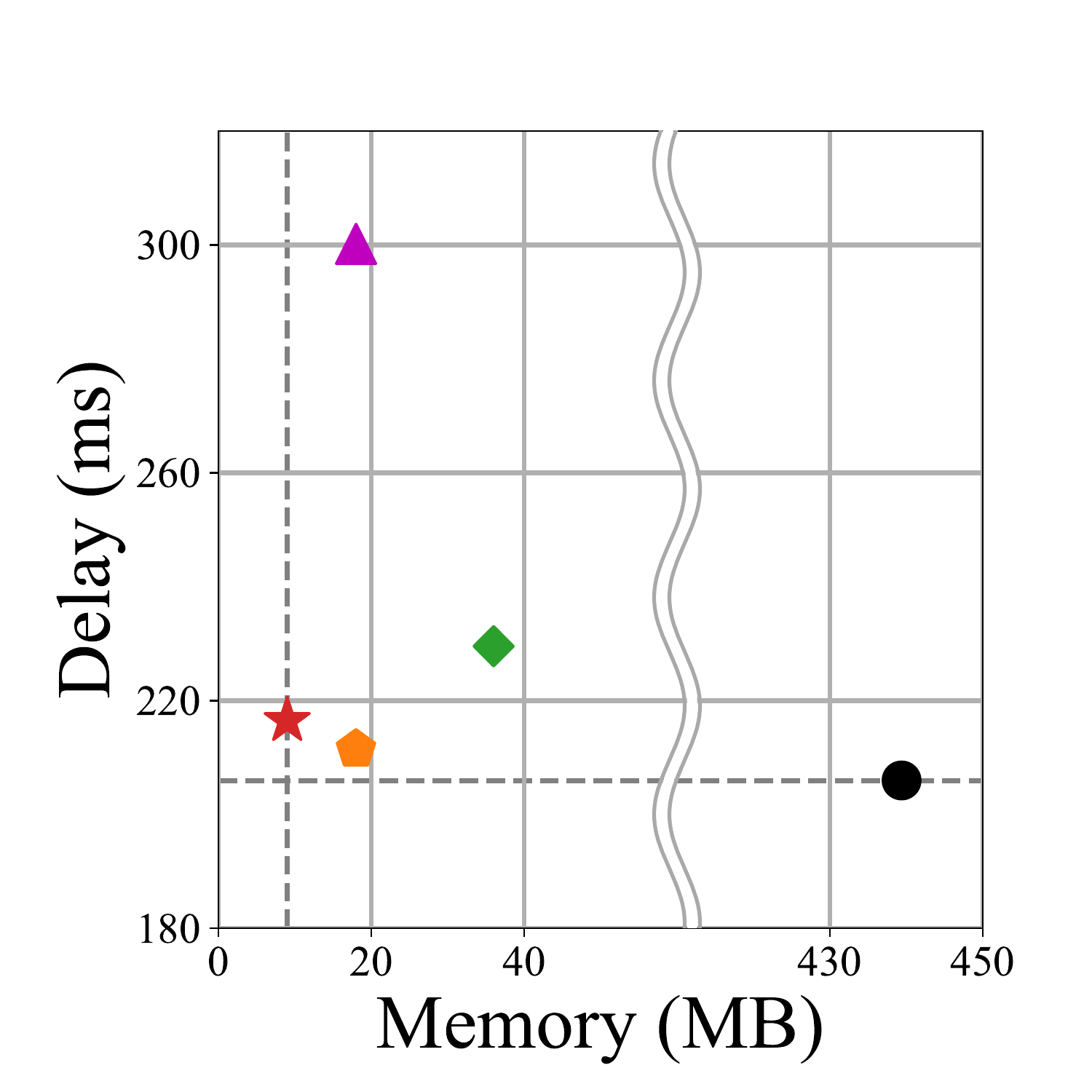}}\hspace{-0.1cm}
    \subfloat[DenseNet.]{\label{fig:densenet_archi}
    \includegraphics[width=0.19\textwidth]{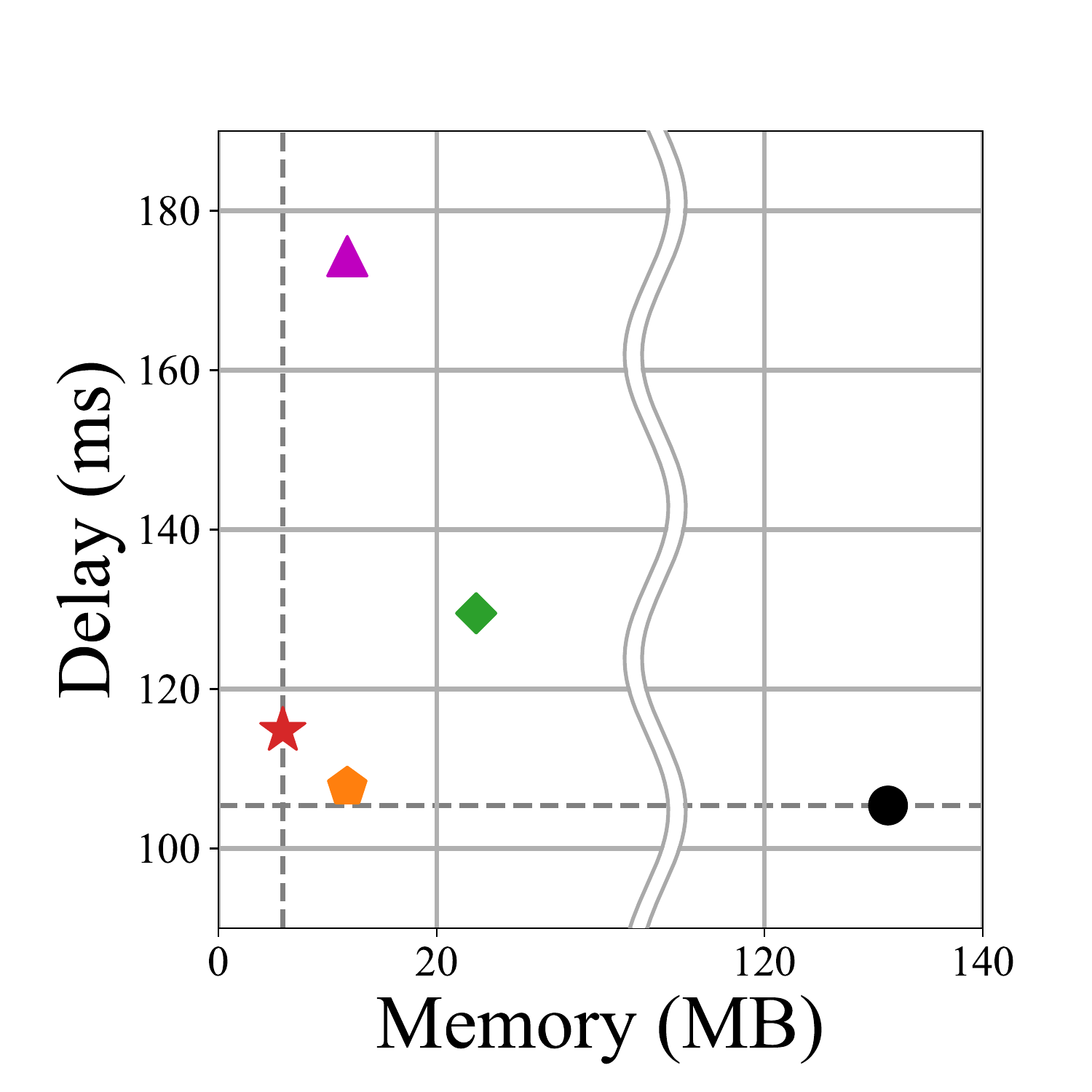}}\hspace{-0.1cm}
\caption{Memory and delay optimization results.}
\label{fig:pipe_opt}
\vspace{-0.7cm}
\end{figure*}

\begin{figure*}
\centering
    \subfloat[\centering{YOLOv3 ($\delta$=1.0~MB).}]{\label{fig:v3_tradeoff}
    \includegraphics[width=0.19\textwidth]{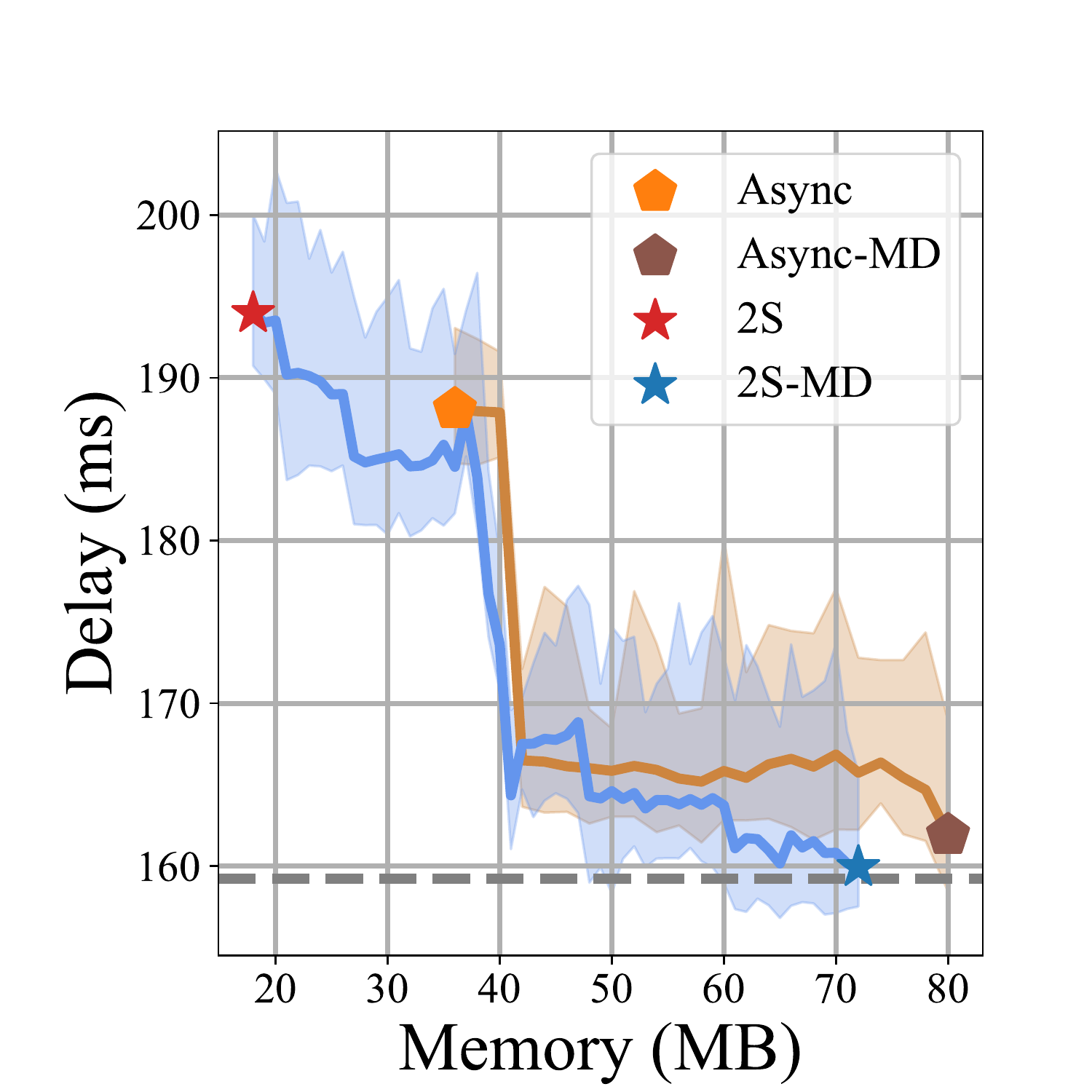}}\hspace{-0.1cm}
    \subfloat[YOLOv4 ($\delta$=1.0~MB).]{\label{fig:v4_tradeoff}
    \includegraphics[width=0.19\textwidth]{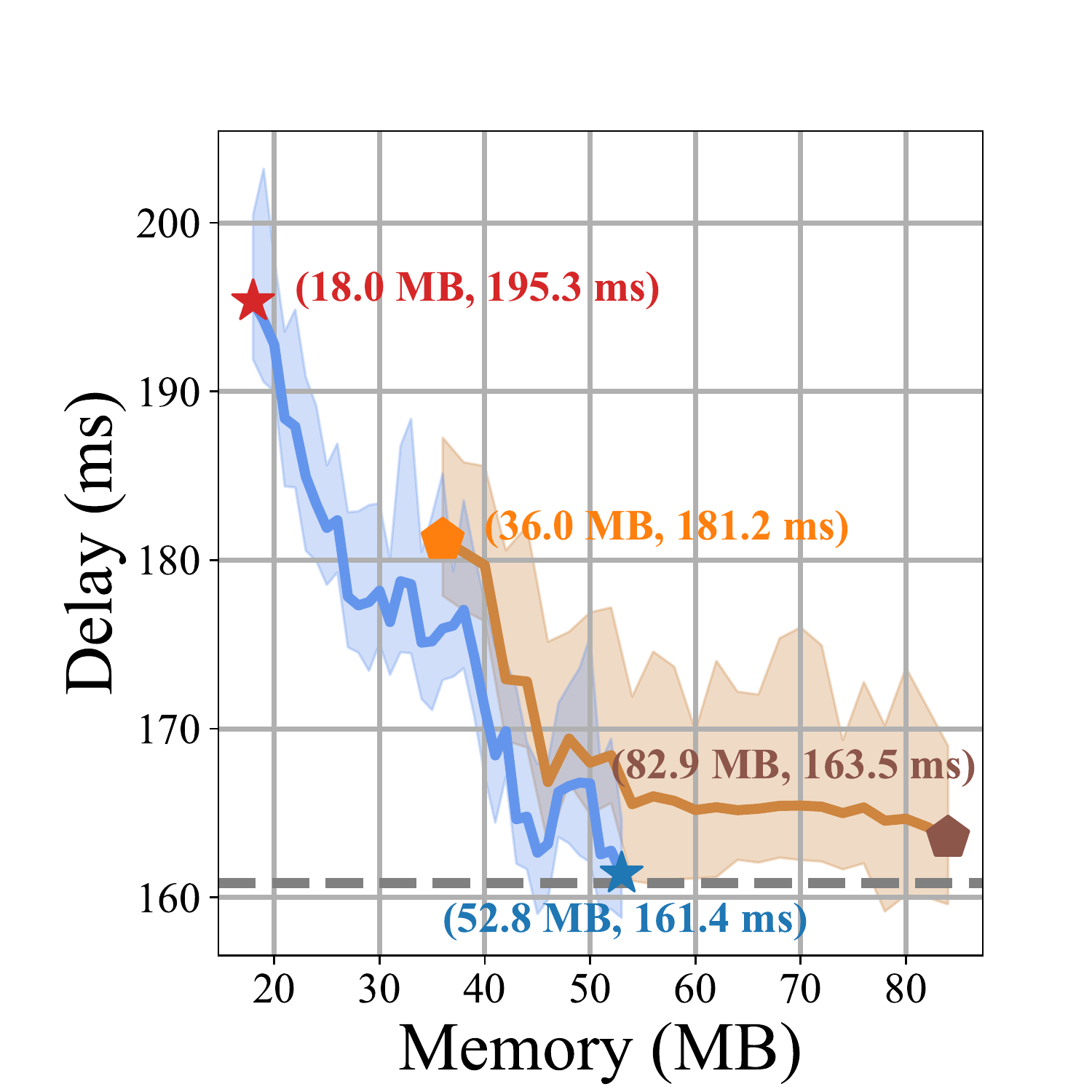}}\hspace{-0.1cm}
    \subfloat[YOLOv4-P6 ($\delta$=1.0~MB).]{\label{fig:v4p6_tradeoff}
    \includegraphics[width=0.19\textwidth]{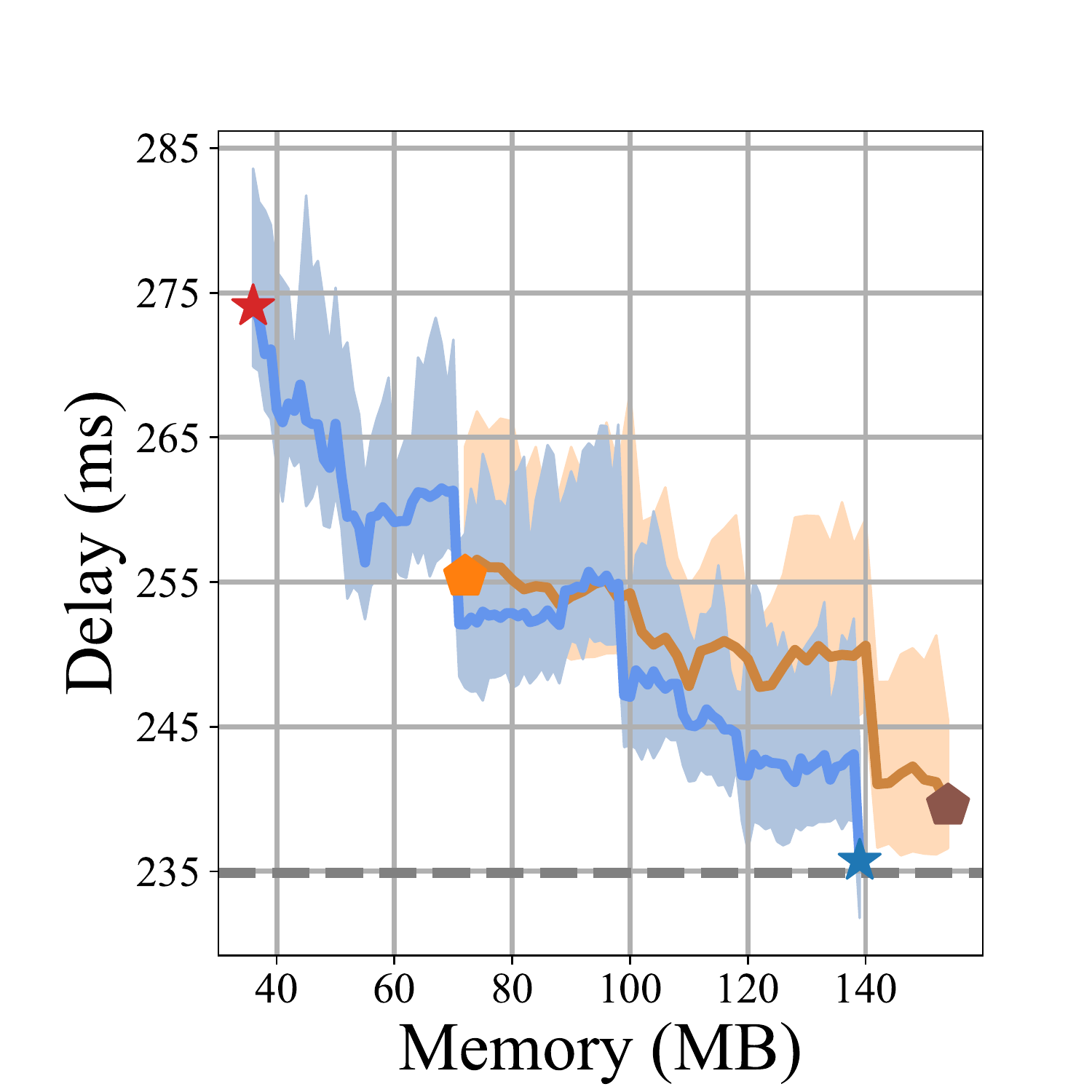}}\hspace{-0.1cm}
    \subfloat[ResNet ($\delta$=0.2~MB).]{\label{fig:resnet_tradeoff}
    \includegraphics[width=0.19\textwidth]{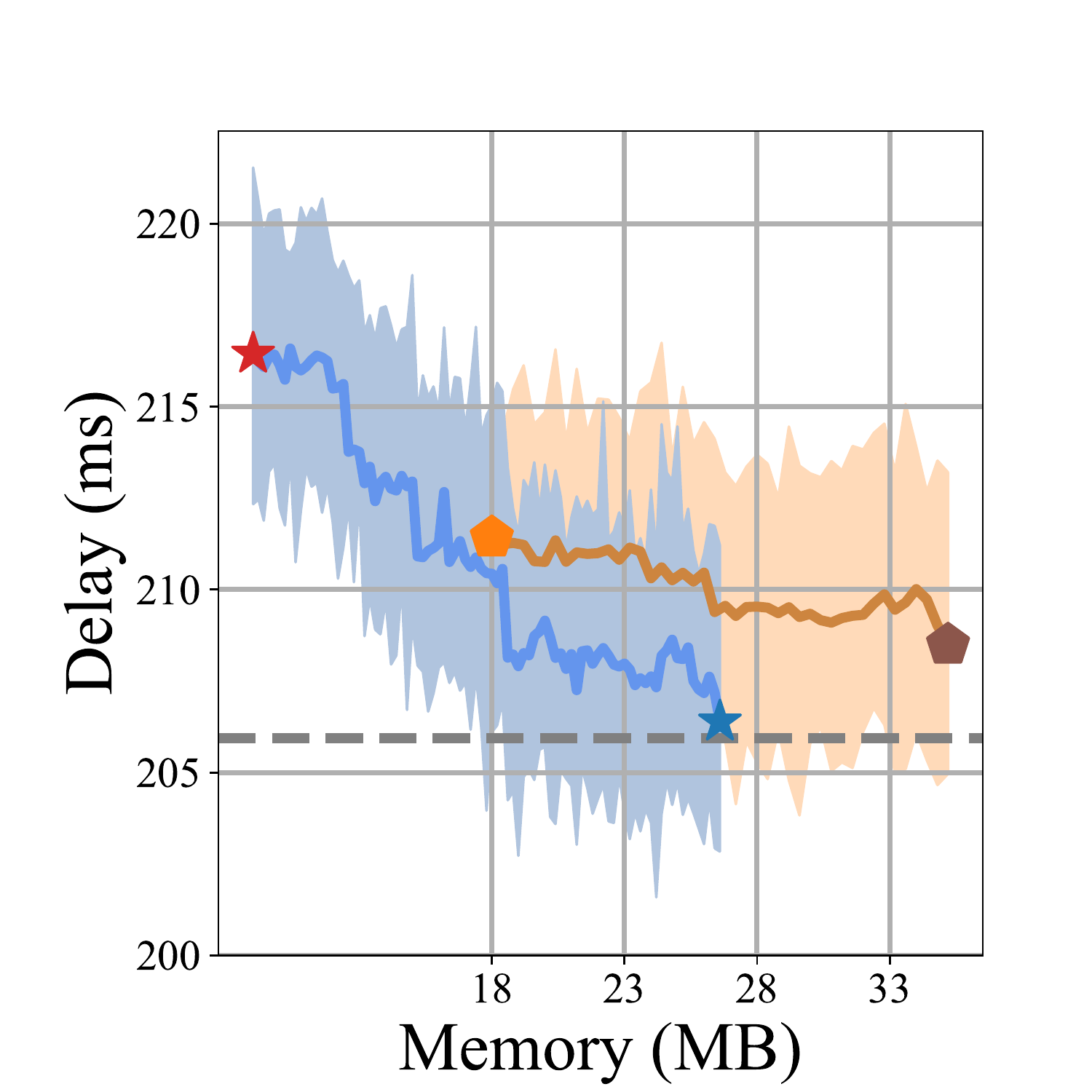}}\hspace{-0.1cm}
    \subfloat[DenseNet ($\delta$=0.2~MB).]{\label{fig:densenet_tradeoff}
    \includegraphics[width=0.19\textwidth]{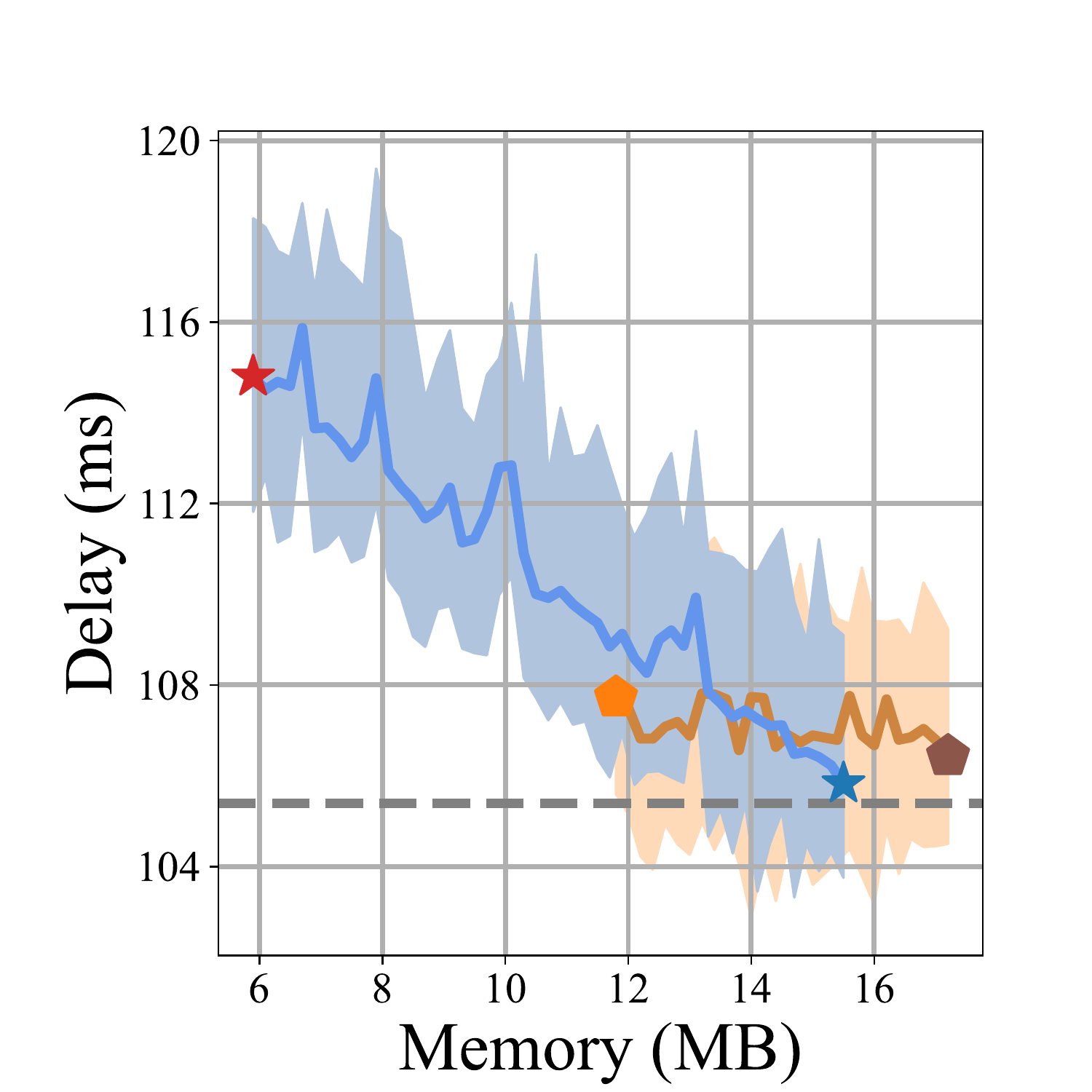}}\hspace{-0.1cm}
\caption{Iterative delay optimization to minimal delay configurations.}
\label{fig:tradeoff}
\end{figure*}
\section{Experiments} \label{sec:experiments}

\subsection{Experimental Setup} \label{sec:impl}

We implemented Demand Layering for Nvidia iGPU platforms.\footnote{The source code is at \url{https://github.com/aveeslab/demand-layering}.} Our experimental platform is Nvidia Jetson AGX Xavier with 16~GB {\em shared} DRAM, an 8-core ARM CPU, and a 512-core integrated Volta GPU. Also, it is equipped with a Samsung 980 PRO NVMe M.2 SSD (1~TB) with its official sequential read performance of 7000~MB/s. As our software platform, we use Nvidia Ubuntu 18.04.6 LTS with CUDA 10.2 and JetPack 4.6.1. As our baseline implementation, we use Darknet~\cite{darknet}, currently available at~\cite{darknet_ab}, which is one of the most famous state-of-the-art DNN frameworks. Darknet is written in C and CUDA, making it portable across various hardware platforms. In Darknet, a DNN model is stored in two separate files. One is a .cfg (text) file that describes the DNN's layer-by-layer structure, including the order and types of layers. The other is a .weights (binary) file, a sequence of parameters that follows the order of layers in the .cfg file. When referring to a DNN model file in this study, it specifically means the .weights file. We use a minutely customized .weights file format to eliminate unnecessary computations in read operations, which is detailed in our GitHub repository.



\begin{table}
\renewcommand{\arraystretch}{1.3}
\caption{DNN models for the evaluation.}
\label{tab:models}
\centering
\begin{tabular}{lcccc}
\toprule
\multirow{2}{*}{\textbf{Model}}  &\textbf{Model}  &  \textbf{Number}   &  \textbf{Maximum} &  \textbf{Default}\\
\multicolumn{1}{c}{} & \textbf{size}  &   \textbf{of layers}   &  \textbf{layer size} &  \textbf{resolution}\\
\midrule
  YOLOv3~\cite{redmon2018yolov3}            & 236.5 MB & 107 & 18.0 MB & 608 x 608\\
  YOLOv4~\cite{bochkovskiy2020yolov4}       & 245.8 MB & 162 & 18.0 MB & 608 x 608\\
  YOLOv4-P6~\cite{wang2021scaledyolov4}     & 487.2 MB & 305 & 36.0 MB & 640 x 640\\
  ResNet152~\cite{he2016deep}                  & 219.6 MB & 206 & 9.0 MB & 608 x 608\\
  DenseNet201~\cite{Huang_2017_CVPR}           & 65.6 MB & 306 & 5.9 MB & 608 x 608\\
\bottomrule
\end{tabular}
\vspace{-0.3cm}
\end{table}

Our current implementation does not support concurrent multi-DNN executions. Thus, all the results are from single-DNN experiments (i.e., one DNN at a time) but with various DNN models. Since real-time scheduling is only meaningful in multitasking environments, we use the stock Linux scheduler rather than its real-time variants. This limitation will be addressed in our future work.

We use the five DNN models as detailed in Table~\ref{tab:models}. Three of them are from the YOLO object detector family. The remaining two are ResNet152 and DenseNet201, which will be referred to as ResNet and DenseNet in short, respectively. YOLOv4-P6 has the largest model size (487.2~MB) and the largest single layer size (36.0~MB). DenseNet is the smallest (65.6~MB) with the largest number of layers (306). In contrast, YOLOv3 has the smallest number of layers (107). The largest layer size ($s^{max}$) varies from 5.9~MB (DenseNet) to 36.0~MB (YOLOv4-P6). Unless otherwise stated, the default input resolution is 608 $\times$ 608 except YOLOv4-P6. As an exception, YOLOv4-P6 has a different default input resolution of 640 $\times$ 640 due to its architectural limitation.

\begin{table*}
\renewcommand{\arraystretch}{1.3}
\caption{Summary of normalized memory and delay optimization results.}
\centering
    \begin{tabular}{lccccccccccccccc}
    \toprule
    \multirow{2}{*}{\textbf{Model}} & 
    \multicolumn{2}{c}{\textbf{Seq}} & \multicolumn{2}{c}{\textbf{Sync}} & \multicolumn{2}{c}{\textbf{Async}} & \multicolumn{2}{c}{\textbf{Async-MD}} & \multicolumn{2}{c}{\textbf{2S}} & \multicolumn{2}{c}{\textbf{2S-MD}}   \\
    
    \multicolumn{1}{c}{}        
    & \textbf{Memory}  &  \textbf{Delay}   &\textbf{Memory}  &   \textbf{Delay}   &\textbf{Memory}  &   \textbf{Delay}   &\textbf{Memory}  &   \textbf{Delay}   &\textbf{Memory}  &   \textbf{Delay}   &\textbf{Memory}  &   \textbf{Delay}    \\
    \midrule
    YOLOv3   & (-92.4\%, & \hspace{-0.5cm} +44.9\%) & (-84.8\%, & \hspace{-0.5cm} +18.9\%) & (-92.4\%, \hspace{-0.5cm} & +18.1\%) & (-83.1\%, \hspace{-0.5cm} & +1.7\%) & (-96.2\%, \hspace{-0.5cm} & +21.8\%) & (-84.8\%, \hspace{-0.5cm} & +0.5\%)\\
    YOLOv4    & (-92.7\%, & \hspace{-0.5cm} +52.1\%) & (-85.4\%, & \hspace{-0.5cm} +23.7\%) & (-92.7\%, \hspace{-0.5cm} & +12.7\%) & (-83.1\%, \hspace{-0.5cm} & +1.7\%) & (-96.3\%, \hspace{-0.5cm} & +21.5\%) & (-89.3\%, \hspace{-0.5cm} & +0.4\%)\\
    YOLOv4-P6 & (-92.6\%, & \hspace{-0.5cm} +70.2\%) & (-85.2\%, & \hspace{-0.5cm} +38.4\%) & (-92.6\%, \hspace{-0.5cm} & +8.7\%)  & (-84.2\%, \hspace{-0.5cm} & +2.0\%) & (-96.3\%, \hspace{-0.5cm} & +16.7\%) & (-85.7\%, \hspace{-0.5cm} & +0.3\%)\\
    ResNet    & (-95.9\%, & \hspace{-0.5cm} +45.8\%) & (-91.8\%, & \hspace{-0.5cm} +11.4\%) & (-95.9\%, \hspace{-0.5cm} & +2.7\%)  & (-92.0\%, \hspace{-0.5cm} & +1.2\%) & (-98.0\%, \hspace{-0.5cm} & +5.1\%)  & (-93.9\%, \hspace{-0.5cm} & +0.2\%)\\
    DenseNet  & (-91.0\%, & \hspace{-0.5cm} +65.4\%) & (-82.0\%, & \hspace{-0.5cm} +22.9\%) & (-91.0\%, \hspace{-0.5cm} & +2.0\%)  & (-86.9\%, \hspace{-0.5cm} & +1.0\%) & (-95.5\%, \hspace{-0.5cm} & +8.9\%)  & (-88.2\%, \hspace{-0.5cm} & +0.4\%)\\
    \bottomrule
    \end{tabular}
\vspace{-0.3cm}
\label{tab:summary}
\end{table*}

\subsection{Evaluation Results}
We compare the following pipeline architectures:
\begin{itemize}
    \item {\bf Pre:} {\em Preloading} architecture.
    \item {\bf Seq:} {\em Sequential} architecture (Section~\ref{sec:sequential}).
    \item {\bf Sync:} 3-stage {\em synchronous} pipeline (Section~\ref{sec:syncpipe}).
    \item {\bf Async:} 3-stage {\em asynchronous} pipeline (Section~\ref{sec:asyncpipe}).
    \item {\bf Async-MD:} {\em Minimal delay} configuration from {\bf Async}.
    \item {\bf 2S:} {\em 2-stage} asynchronous pipeline (Section~\ref{sec:bleeding}).
    \item {\bf 2S-MD:} {\em Minimal delay} configuration from {\bf 2S}.
\end{itemize}
For the iterative optimization for finding minimal delay configurations, refer to Section~\ref{sec:tradeoff}.

Fig.~\ref{fig:pipe_opt} shows the memory requirement (x-axis) and inference delay (y-axis) of the considered pipeline architectures. Every data point in the figure is an average of 1000 iterations. In particular, Fig.~\ref{fig:v4_archi} shows the results for YOLOv4, where {\bf Pre} requires a huge amount of memory (491.6~MB) with YOLOv4's optimal delay (160.8~ms). With {\bf Seq}, the memory requirement is significantly dropped (to 36.0~MB) at the cost of a delay increase (to 244.7~ms). By applying {\bf Sync}, the delay is somewhat reduced (to 198.9~ms), however, at the cost of a memory increase (to 72.0~MB) due to the double-buffering scheme. With {\bf Async}, the memory requirement goes back (to 36.0~MB) and the delay is preferably reduced (to 181.2~ms). By applying {\bf 2S}, the memory requirement is finally optimal, which is a single layer size ($s^{max}$ = 18.0~MB), with a slight delay increase (to 195.3~ms). The other DNNs show similar patterns.

Fig.~\ref{fig:tradeoff} shows the iterative delay optimization process and the resulting minimal delay configurations. This optimization method is only applicable to asynchronous pipeline architectures (i.e., {\bf Async} and {\bf 2S}), where we can freely adjust the buffer size above the minimum requirement. In the figure, the solid lines depict average delays as gradually increasing buffer sizes, while the colored areas illustrate min-max delay ranges. In particular, Fig.~\ref{fig:v4_tradeoff} shows the results of YOLOv4. During the optimization, {\bf Async} begins at (36.0~MB, 181.2~ms) and optimized to (82.9~MB, 163.5~ms). In comparison, {\bf 2S} begins at (18.0~MB, 195.3~ms) and optimized to (52.8~MB, 161.4~ms). In both cases, the delay overhead is significantly minimized, closely approaching the optimal delay (160.8~ms) depicted by a dashed line.

\begin{figure*}
    \centering
    \subfloat[YOLOv3.]{
    \label{fig:v3_DOpt}
    \includegraphics[width=0.2\textwidth]{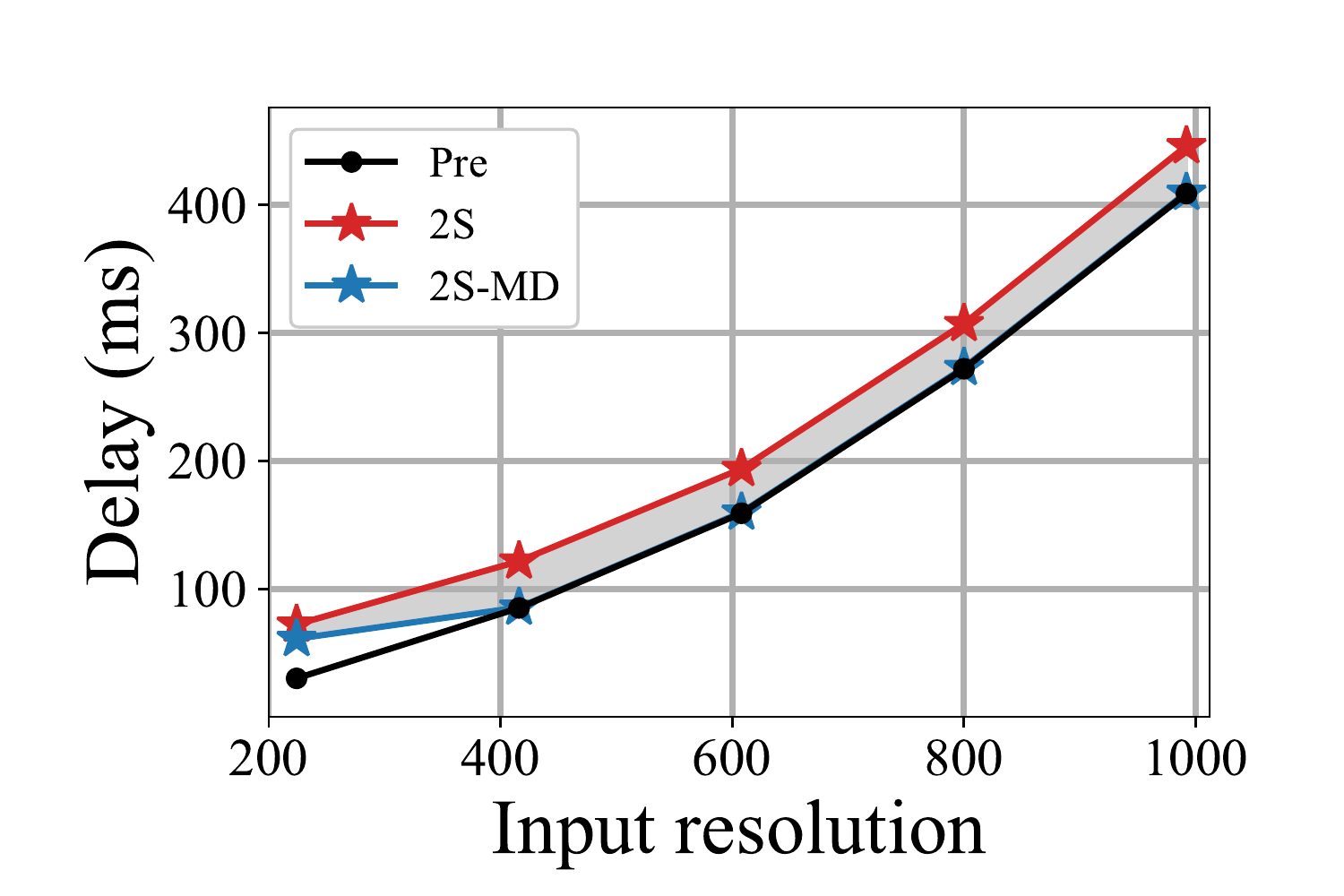}}\hspace{-0.3cm}
    \subfloat[YOLOv4.]{
    \label{fig:v4_DOpt}\includegraphics[width=0.2\textwidth]{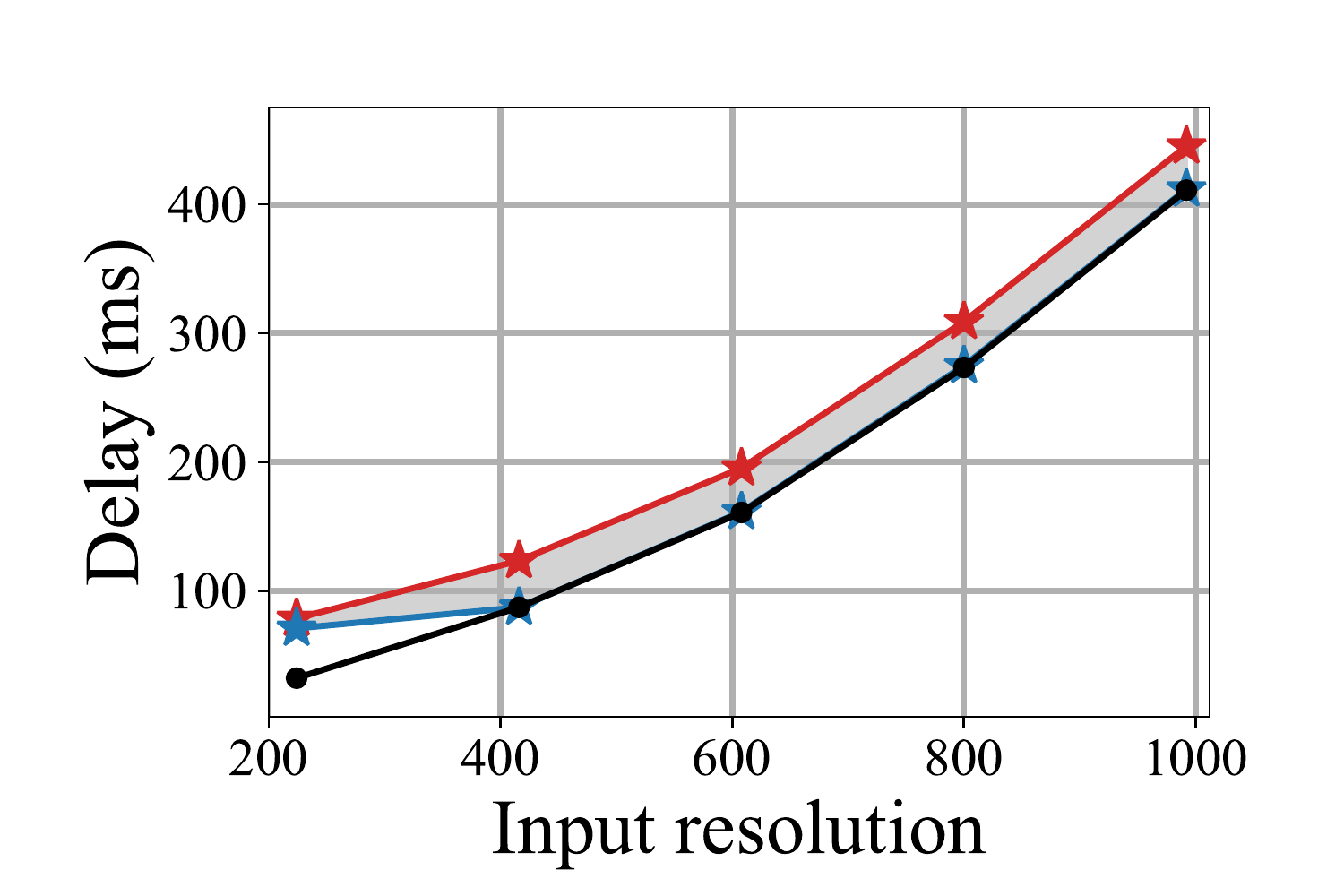}}\hspace{-0.3cm}
    \subfloat[YOLOv4-P6.]{
    \label{fig:v4p6_DOpt}\includegraphics[width=0.2\textwidth]{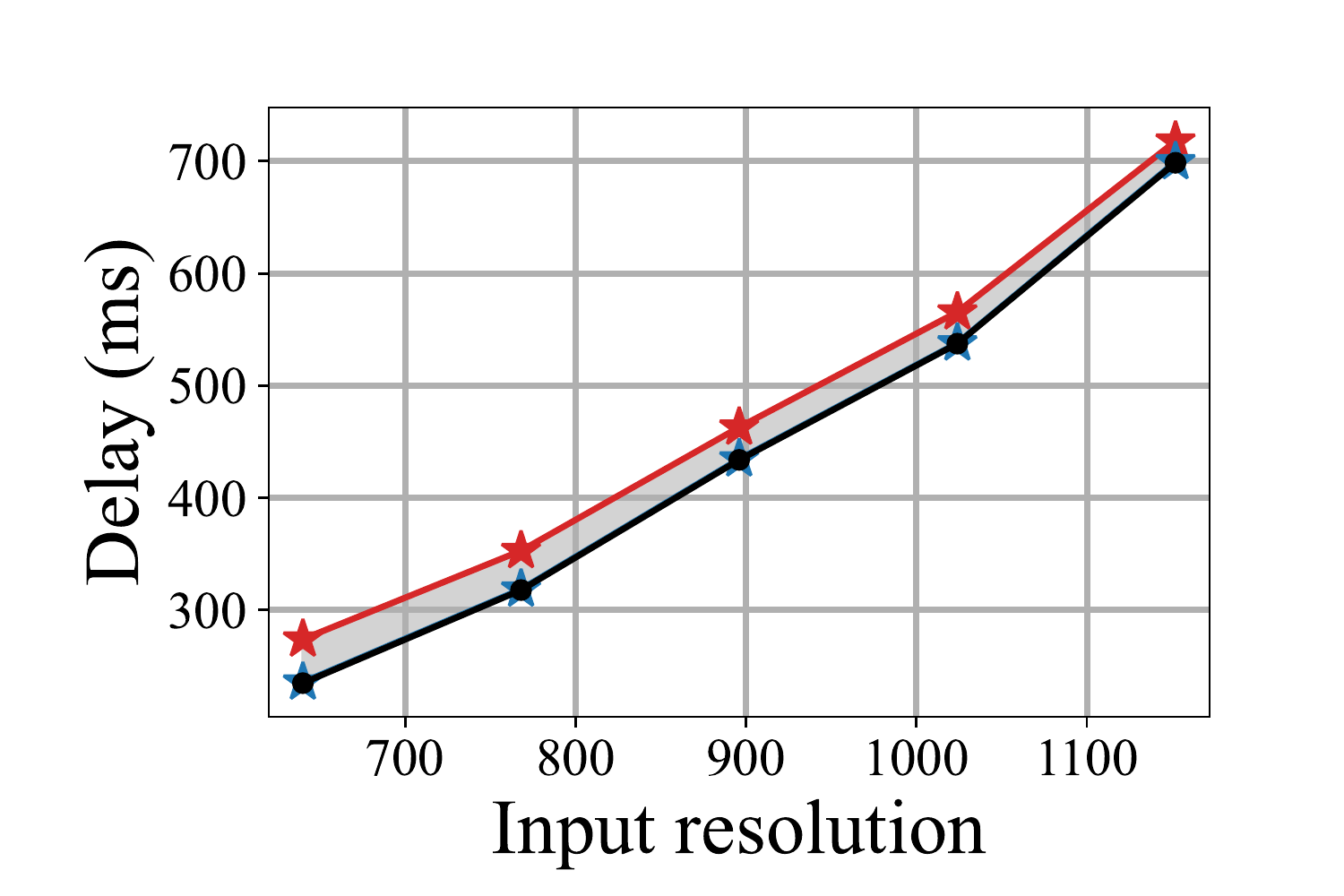}}\hspace{-0.3cm}
    \subfloat[ResNet.]{
    \label{fig:resnet_DOpt}\includegraphics[width=0.2\textwidth]{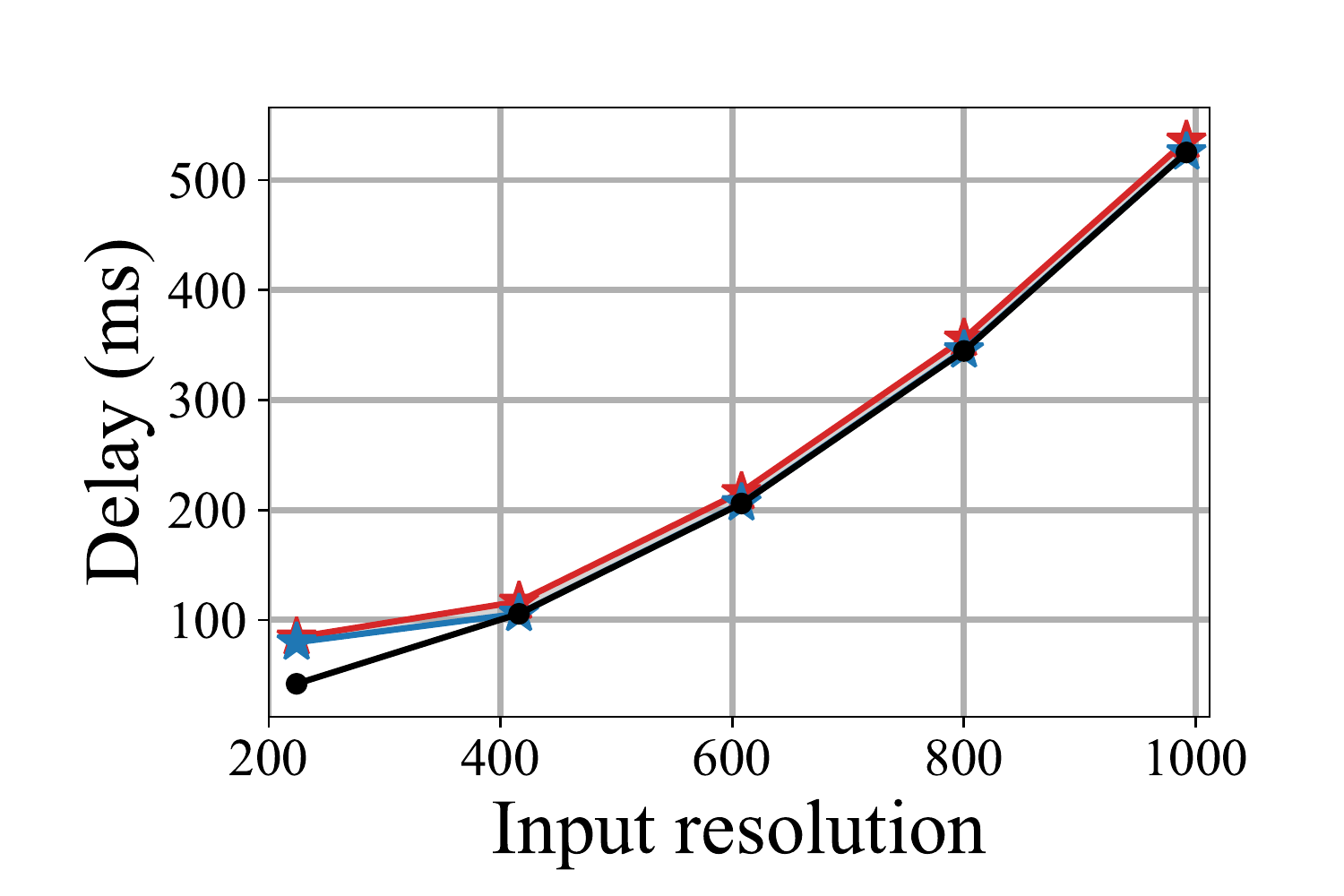}}\hspace{-0.3cm}
    \subfloat[DenseNet.]{
    \label{fig:densenet_DOpt}\includegraphics[width=0.2\textwidth]{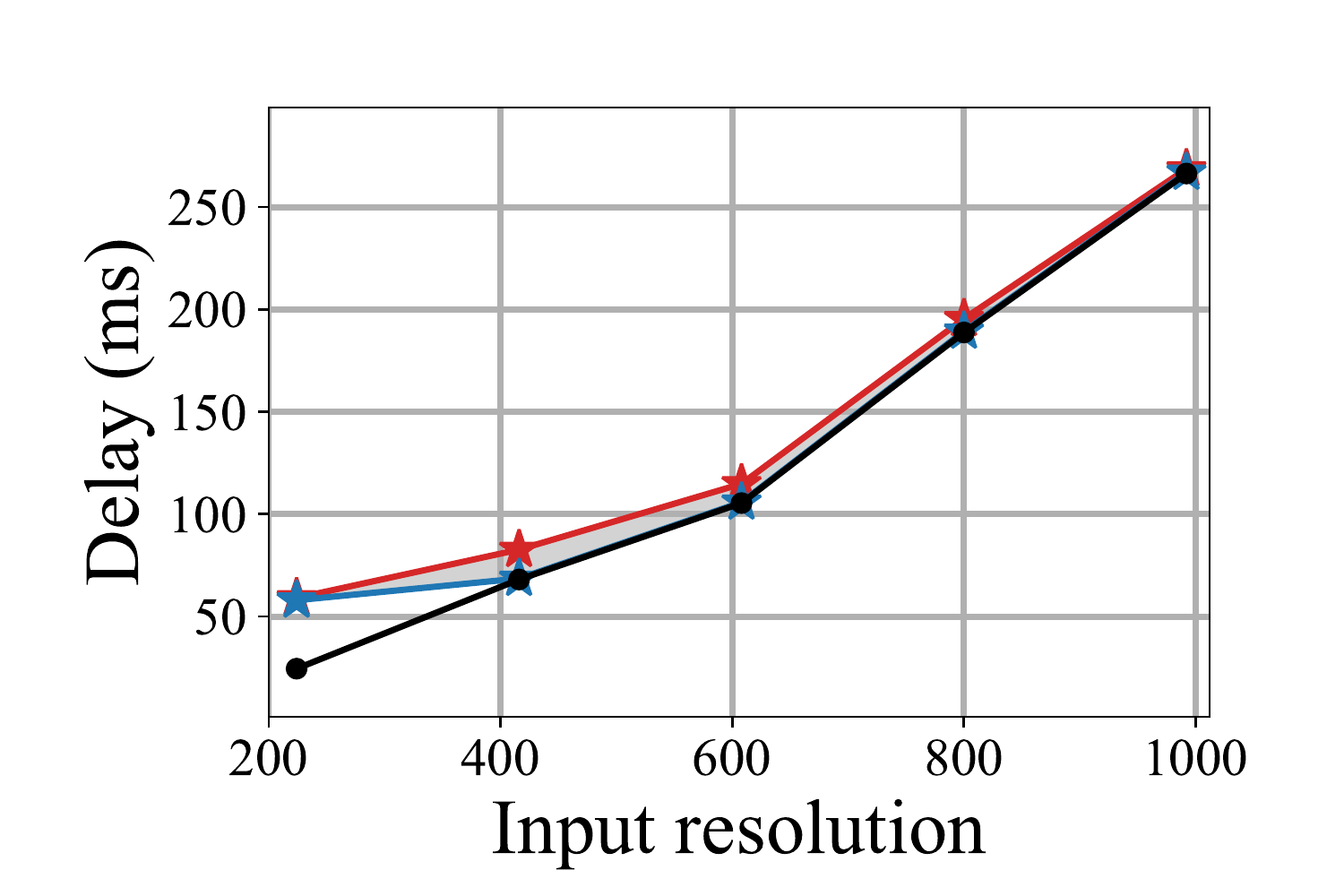}}\hspace{-0.3cm}
\caption{Delay optimization results with varying input image resolutions.}
\label{fig:NN_DOpt}
\end{figure*}

\begin{figure*}
    \centering
    \subfloat[YOLOv3.]{
    \label{fig:v3_MOpt}
    \includegraphics[width=0.2\textwidth]{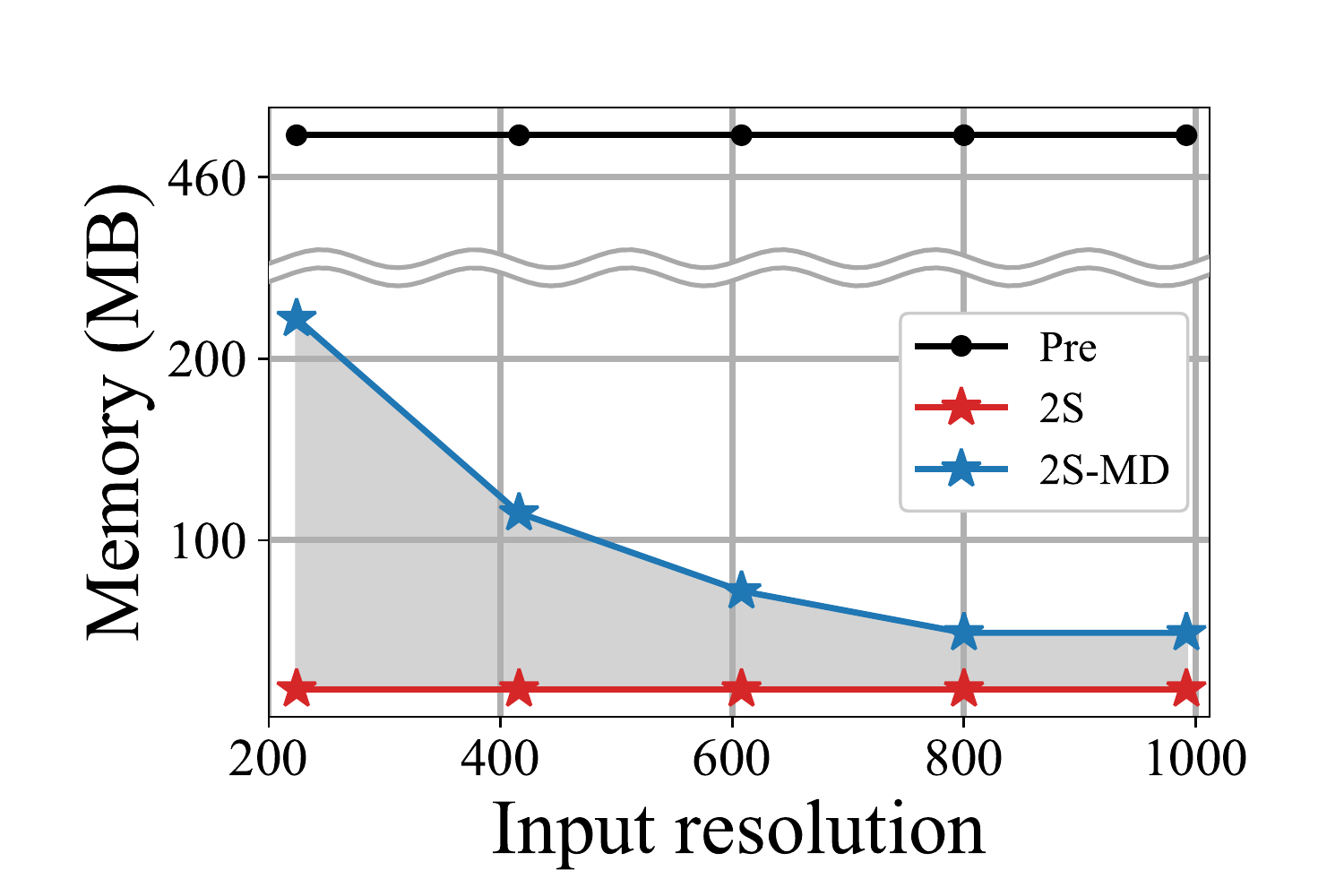}}\hspace{-0.3cm}
    \subfloat[YOLOv4.]{
    \label{fig:v4_MOpt}\includegraphics[width=0.2\textwidth]{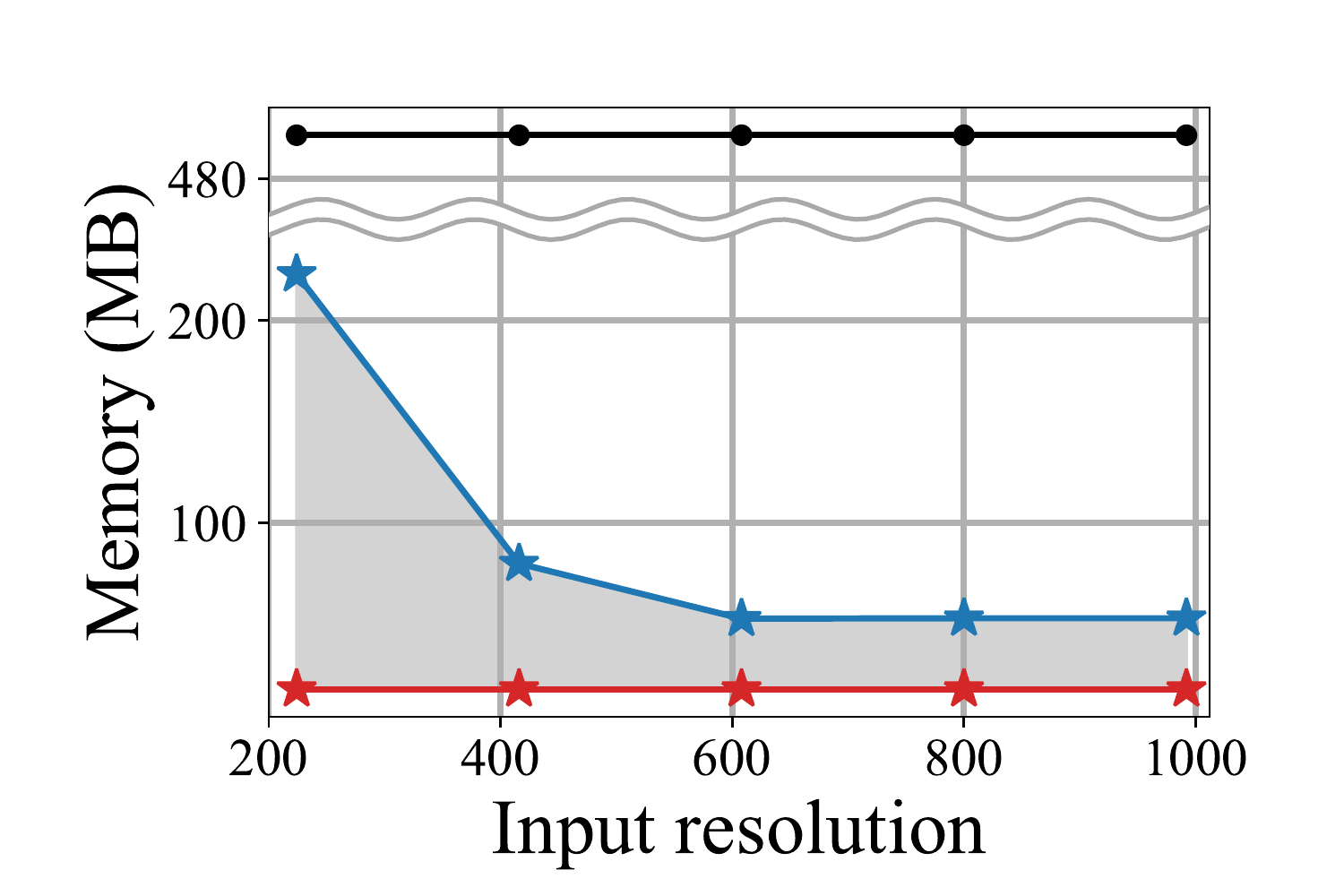}}\hspace{-0.3cm}
    \subfloat[YOLOv4-P6.]{
    \label{fig:v4p6_MOpt}\includegraphics[width=0.2\textwidth]{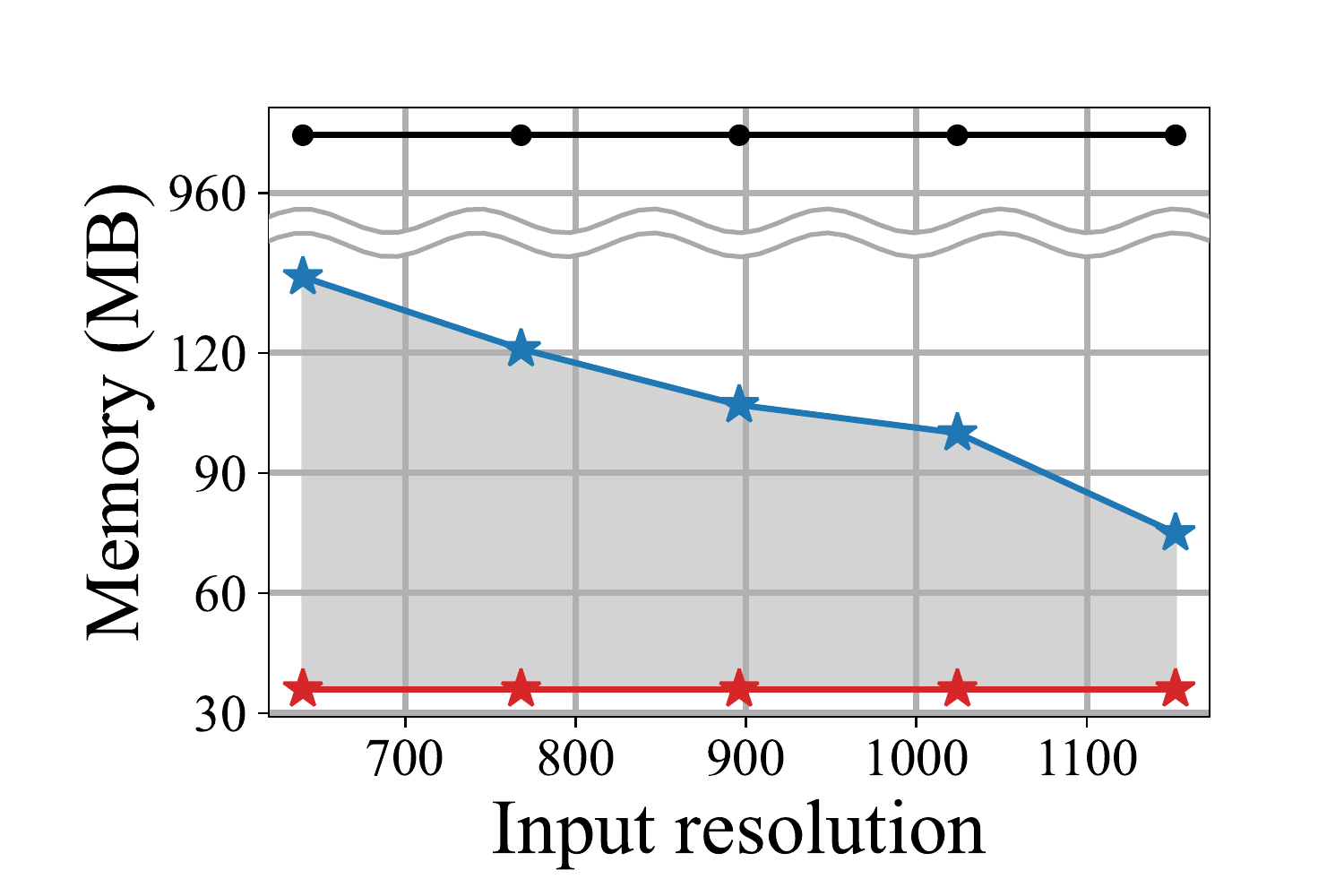}}\hspace{-0.3cm}
    \subfloat[ResNet.]{
    \label{fig:resnet_MOpt}\includegraphics[width=0.2\textwidth]{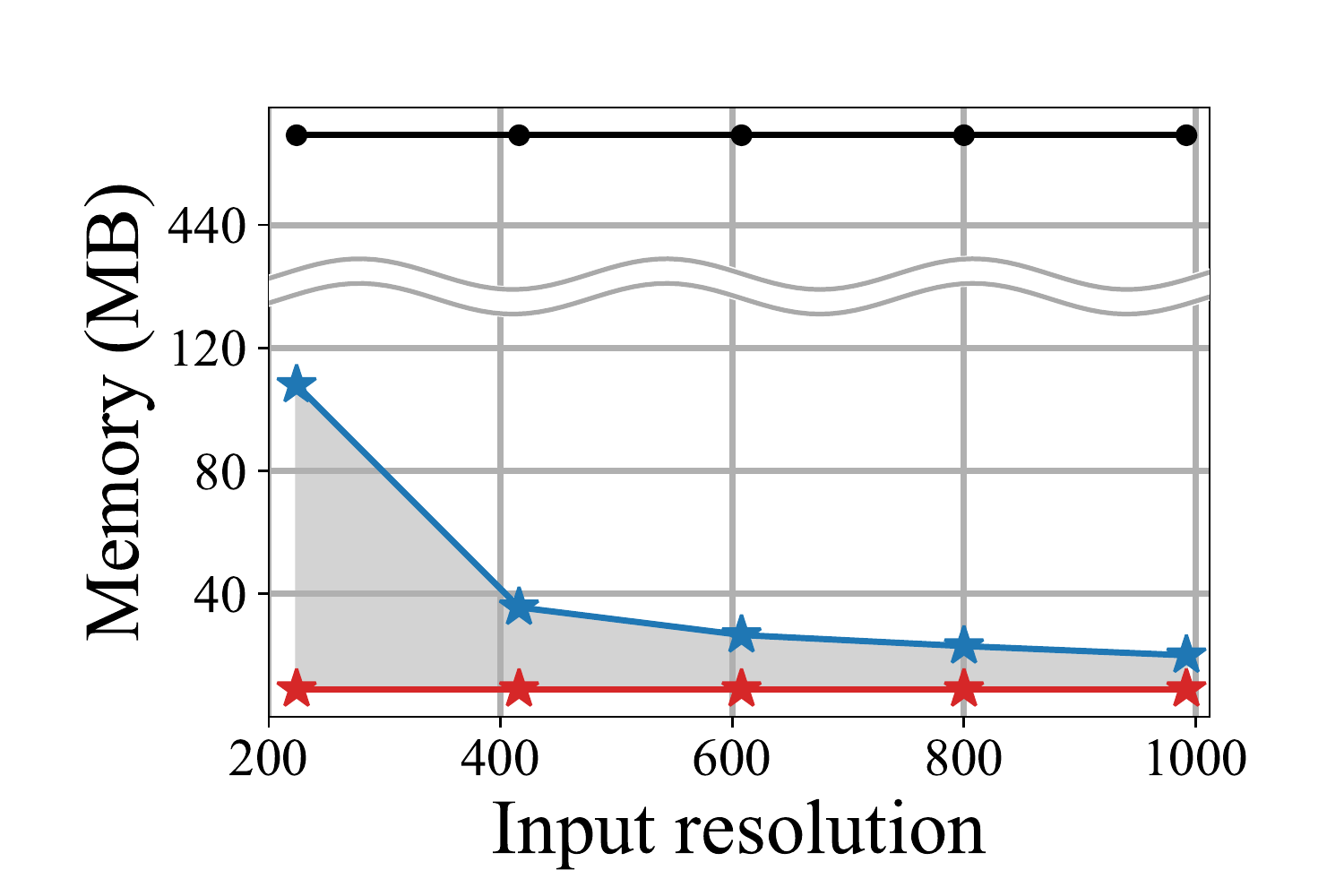}}\hspace{-0.3cm}
    \subfloat[DenseNet.]{
    \label{fig:densenet_MOpt}\includegraphics[width=0.2\textwidth]{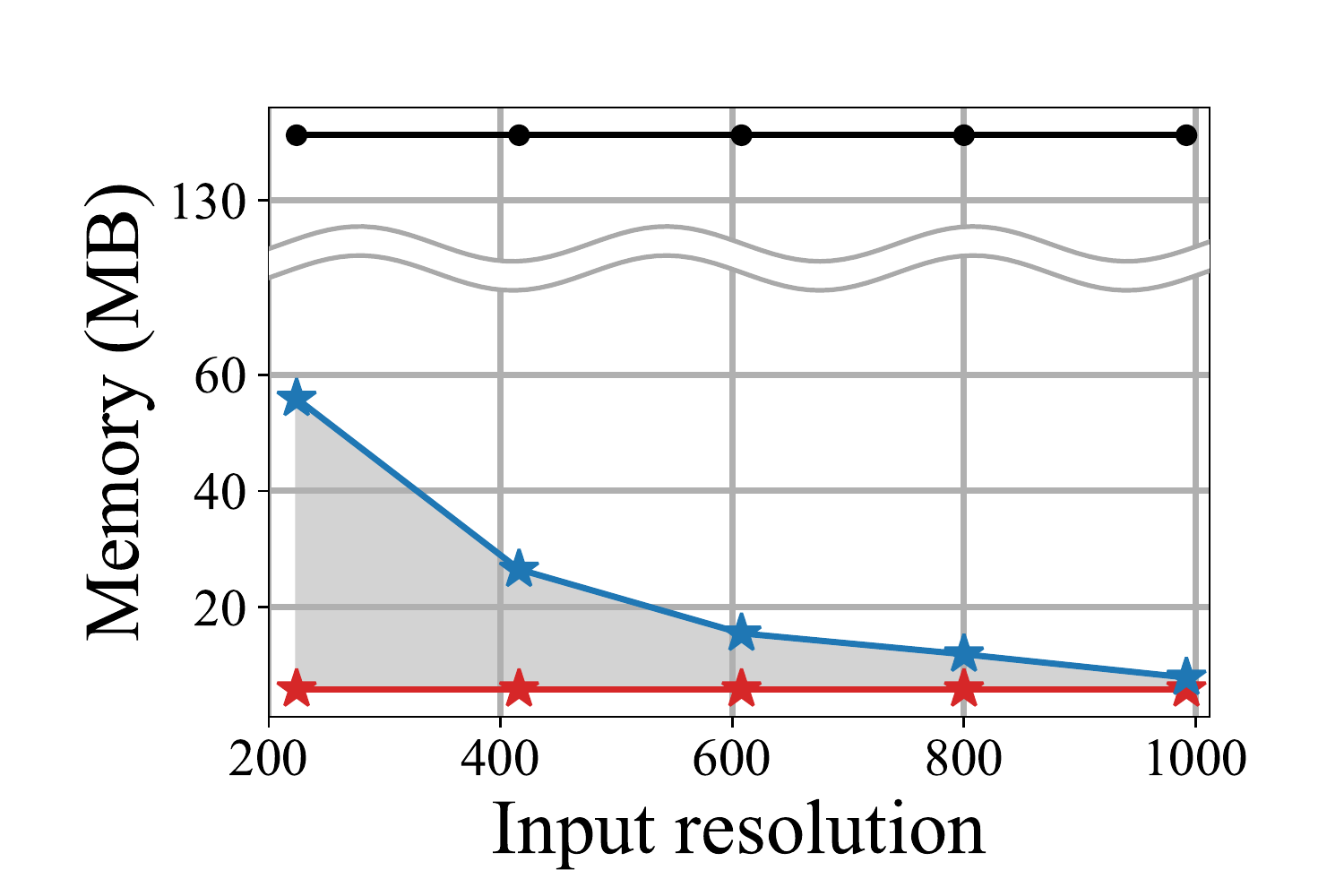}}\hspace{-0.3cm}
\caption{Memory optimization results with varying input image resolutions.}
\label{fig:NN_MOpt}
\end{figure*}

\begin{figure}
    \centering
    \includegraphics[width=0.45\textwidth]{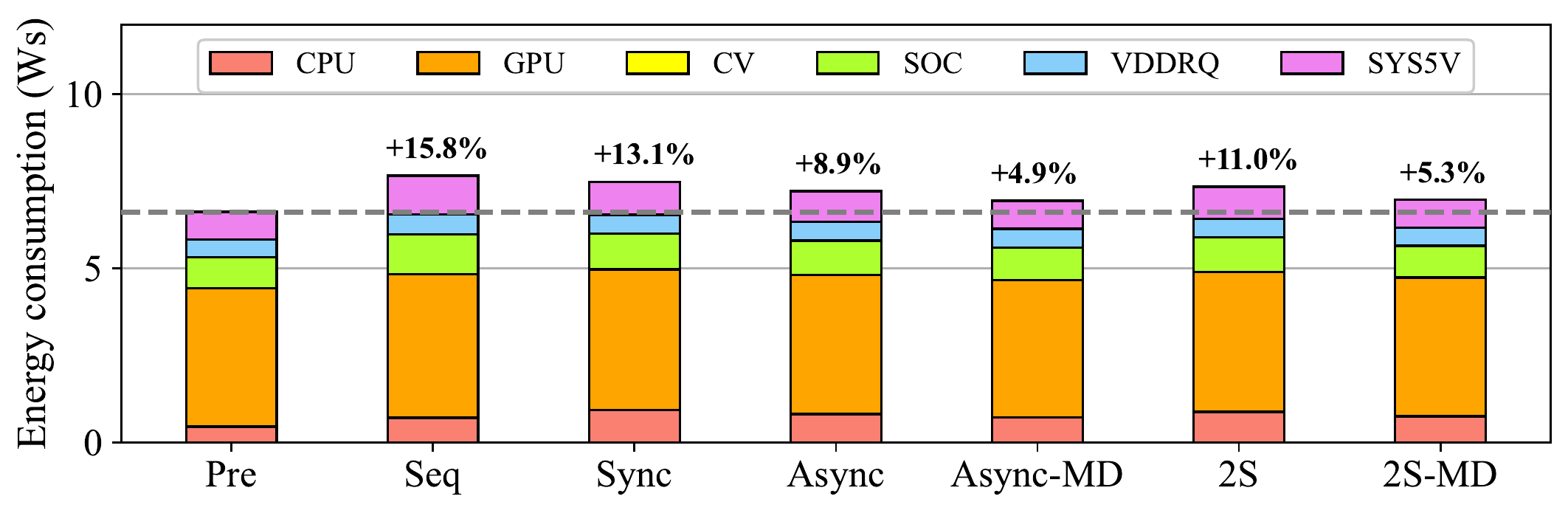}
    \caption{Comparison of per-inference energy consumption of YOLOv4.}
    \label{fig:energy}
\end{figure}

Table~\ref{tab:summary} summarizes the normalized memory reduction and delay overhead from the baseline architecture {\bf Pre} that has the maximal memory requirement with the optimal delay. In general, {\bf 2S} shows the most significant memory reductions (-96.5\% on average), while {\bf 2S-MD} shows minimal delays close to the zero-delay overhead with still significant memory reductions (-88.4\% on average). However, {\bf Async} and {\bf Async-MD} also show comparable optimization results for minimal memory and minimal delay configurations, respectively. Thus, in systems without the support of zero-copy memory, our guide is to use one of the optimization configurations between {\bf Async} and {\bf Async-MD} in Fig.~\ref{fig:tradeoff}. In systems supporting zero-copy memory, using the results between {\bf 2S} and {\bf 2S-MD} in Fig.~\ref{fig:tradeoff} provides even better results. However, beware that it is still based on a bleeding-edge technology.

The input image resolution significantly impacts kernel times and the overall optimization results. Fig.~\ref{fig:NN_DOpt} shows its impact on delays, where the higher input resolution generally incurs the longer delay. By comparing the five DNNs, {\bf 2S-MD} in ResNet and DenseNet particularly show near-optimal delays close to {\bf Pre}. Since ResNet and DenseNet are well balanced with similarly sized layers, they are favorable for pipelining. Even so, when the input resolution gets too small, the gap between {\bf 2S-MD} and {\bf Pre} visibly arises due to the reduced optimization opportunity by too short kernel times. By looking at how much delay is reduced from {\bf 2S} to {\bf 2S-MD} (the grey area) for each DNN, all the DNNs except YOLOv4-P6 fail to reach the near-optimal delay when the input resolution gets too small (i.e., 224 $\times$ 224). Even for the YOLOv4-P6 case, it is not a complete experiment because we cannot lower the input resolution under a certain level due to its technical limitation. Besides, Fig.~\ref{fig:NN_MOpt} shows how much memory is incremented during the iterative optimization from {\bf 2S} to {\bf 2S-MD}, depicted by the grey area. The decreasing trend indicates that the memory cost spent for reducing the delay is significantly lower with higher input resolutions. In other words, the trading cost is dependent on the input resolution. With higher input resolutions, we can easily reduce delays by increasing a small amount of memory.


Fig.~\ref{fig:energy} compares the per-inference energy consumption for various architectures, broken down by the hardware components of interest. For that, our target platform has two 3-channel INA3221 power monitors so that we can extract the energy consumption of the six hardware components through the /proc interface. In the graph, SOC denotes the Xavier SoC portion, excluding the CPU, GPU, and vision accelerator (CV) portions; VDDRQ denotes the external DRAM portion; SYS5V represents the remaining portion including attached I/O devices. It is certain that {\bf Pre} exhibits the minimal energy consumption. By using Demand Layering, the energy consumption is somewhat increased due to the additional read and copy operations during the inference phase. Also, we need a scheduling thread that did not exist in {\bf Pre}, making a CPU core busy. Although the total amount of work (i.e., read, copy, and kernel operations) is the same across the remaining six Demand Layering architectures, there is a slight variance among them that has a strong correlation with their inference delays, because longer delays incur more baseline energy consumption during an inference. However, the additional energy consumption by Demand Layering is not significant.

\section{Related Work} \label{sec:related}


{\bf DNN frameworks.} There are various DNN frameworks~\cite{abadi2016tensorflow, jia2014caffe, caffe2, paszke2019pytorch, chen2015mxnet, darknet}, supporting both training and inference tasks. Besides, there are frameworks specialized in inference such as TensorRT~\cite{tensorrt}, TensorFlow Lite~\cite{warden2019tinyml},  TensorFlow Lite Micro~\cite{david2021tensorflow}, and TinyEngine~\cite{lin2020mcunet}. To the best of our knowledge, the above frameworks commonly employ the preloading architecture, incurring significant memory overhead that is our concern. Among them, we use Darknet as our baseline implementation.


{\bf Model optimization.} Most DNN frameworks have their own model file formats, making it challenging to exchange models across frameworks. There are standardization efforts to alleviate this problem, such as ONNX~\cite{onnx}. Before deploying models, it is a usual practice to optimize the model such that it can run faster with less memory. For that, various model compression techniques are used, such as pruning and quantization~\cite{han2015deep, he2018amc, liu2017learning}. Additionally, there are compiler frameworks~\cite{li2020deep, rotem2018glow, chen2018tvm, xla} that optimize given models to target hardware platforms, including FPGAs and custom ASICs. Our approach does not compete with the above optimization techniques. In contrast, our method accepts such optimized models and can still significantly reduce memory usage without modifying the model itself.

{\bf DNN memory optimization.} To minimize the memory usage of DNN inference systems, TensorFlow Lite reuses activation buffers that store intermediate results between layers~\cite{pisarchyk2020efficient}. \cite{lin2020mcunet} and \cite{lin2021mcunetv2} optimize memory usage for MCU hardware by combining neural architecture optimizations and cross-layer patch-based computations. \cite{liberis2019neural} reduces peak memory usage by reordering layer executions. \cite{alwani2016fused} reduces off-chip memory usage by fusing multiple CNN layers to utilize on-chip memory efficiently. \cite{miao2021enabling} reduces memory usage by swapping out tensors to external flash storages. \cite{goetschalckx2019breaking} reduces on-chip memory usage and I/O bandwidth by doing calculations across layers. Most of the above studies reduce memory usage by activation buffers. In contrast, we minimize memory usage by model parameters. Also, they do not consider GPU-based systems.

{\bf GPU memory optimization.} Due to the scarcity of GPU memory, there have been many studies\cite{rhu2016vdnn, sekiyama2018profile, yang2020efficient, shah2020memory, peng2020capuchin, narayanan2021memory} that minimize GPU memory usage when training large DNNs, which is not this study's concern. In contrast, we focus on minimizing a DNN inference system's GPU memory usage to deploy large DNNs on embedded systems with limited memory. With the same motivation, SwapAdvisor~\cite{huang2020swapadvisor} provides a general method that utilizes inexpensive CPU memory as a swap device of scarce GPU memory. In dGPU systems, it is promising. However, reducing GPU memory at the cost of increased CPU memory is not meaningful in iGPU systems. Refer to Section~\ref{sec:integrated} for more information. In \cite{bateni2020co}, a memory/performance co-optimization framework for multitasking GPU applications is proposed, where three different GPU memory types are empirically analyzed. Although it does not utilize the layer-by-layer execution of DNNs, the analysis results gave us great insight.

{\bf Layer-by-layer DNN execution.} There are studies exploiting the layer-by-layer execution of DNNs. LaLaRAND~\cite{kang2021lalarand} improves the schedulability of real-time DNN tasks by optimally allocating and scheduling individual layers to CPU and GPU. PipeSwitch~\cite{bai2020pipeswitch} employs an idea of layer-by-layer DNN parameter loading and execution, which is similar to ours. However, PipeSwitch does not consider the read stage, and its goal is to enable fast context switching for multi-DNN systems, which is not related to the memory usage issue. MASA~\cite{cox2021masa, cox2022memory} also uses a similar idea that resembles our layer-by-layer loading and execution. However, MASA assumes only CPUs without considering GPUs. TASO~\cite{wen2020taso} is also based on a similar idea. However, it only supports convolutional layers by optimally selecting layer implementations (e.g., im2col and Winograd). In contrast, our approach generally supports any DNN layers, including fully connected layers as well as any unseen custom layers.

{\bf PREM architecture.} Our method has some technical similarities with the PRedictable Execution Model (PREM) architecture~\cite{pellizzoni2011predictable}, where each task is split into a number of I/O phases, memory phases, and execution phases, and multiple such tasks are coscheduled to avoid the shared resource (e.g., cache, memory, and bus) contention. In \cite{wasly2014hiding, tabish2016real, tabish2019real, casini2020predictable}, the local memory (e.g., scratchpad memory) is partitioned into two regions such that a memory phase (by DMA controller) and an execution phase (by CPU) can run in a pipelined parallel manner. They are conceptually similar to our read, copy, and kernel pipeline stages and the double-buffering scheme in the synchronous pipeline architecture. It is interesting to note that the techniques developed for the conventional real-time task model can be revisited for emerging applications (e.g., real-time DNN inference). PREM's primary objective is to eliminate the shared resource contention for predictable systems. Thus, they do not use the asynchronous pipeline technique, which may break the predictability. In contrast, we primarily focus on the memory usage, where the asynchronous pipeline is generally a better solution.

\section{Conclusion} \label{sec:conclusion}
This study presents Demand Layering that minimizes the DNN inference system's memory usage for model parameters by loading and executing DNN layers in a layer-by-layer manner. To minimize the delay overhead, we designed a pipeline architecture where the read, copy, and kernel operations run in parallel. It is further enhanced by employing the asynchronous pipeline and zero-copy memory. As a result, we can reduce the memory usage by 96.5\% with just 14.8\% delay overhead. Also, we can achieve near-zero delay overhead with a still significant 88.4\% memory reduction by exploiting the memory-delay tradeoff.

In the future, we plan to develop a deterministic delay analysis method such that a given model's inference delay can be safely bounded from the model configuration. Also, we plan to develop a scheduling algorithm for concurrent DNNs. Ultimately, our final goal is to make a deterministic multi-DNN inference architecture with minimized memory usage.

\section*{Acknowledgment}
This work was supported partially by the BK21 Four Program (5199990814084) of the National Research Foundation of Korea (NRF) funded by the Ministry of Education, Korea, partially by the National Research Foundation of Korea (NRF) grant funded by the Korean government (MSIT) (2022R1A2C1013197), and partially by the NSF grant CCF-1704859. Mingoo Ji's work was done while at Kookmin University. J.-C. Kim is the corresponding author of this paper.

\bibliographystyle{IEEEtran}
\bibliography{demand}

\end{document}